\documentclass{article}

\usepackage[preprint]{neurips_2025}

\usepackage[utf8]{inputenc} % allow utf-8 input
\usepackage[T1]{fontenc}    % use 8-bit T1 fonts
\usepackage{hyperref}       % hyperlinks
\usepackage{url}            % simple URL typesetting
\usepackage{booktabs}       % professional-quality tables
\usepackage{amsfonts}       % blackboard math symbols
\usepackage{nicefrac}       % compact symbols for 1/2, etc.
\usepackage{microtype}      % microtypography
\usepackage{xcolor}         % colors
\usepackage{amsfonts}       % blackboard math symbols
\usepackage{amsmath}        % for \text and math environments
\usepackage{graphicx} % 用于 \resizebox
\usepackage{booktabs} % 用于 \toprule, \midrule, \bottomrule
\usepackage{multirow} % 用于 \multirow
\usepackage{makecell}
\usepackage{subcaption}
\DeclareMathOperator{\Warp}{Warp}
\DeclareMathOperator{\clip}{clip}     % clip operator <--- 已添加
 
\title{Post-Training Quantization for Video Matting}

\author{
  \begin{tabular}{c} % 外层表格，确保整体居中
    Tianrui Zhu\textsuperscript{1} \quad \quad
    Houyuan Chen\textsuperscript{1} \quad \quad
    Ruihao Gong\textsuperscript{2} \\
    Michele Magno\textsuperscript{3} \quad \quad
    Haotong Qin\textsuperscript{3}\thanks{Corresponding authors.} \quad \quad
    Kai Zhang\textsuperscript{1}\footnotemark[1]
  \end{tabular}
  \\
  \textsuperscript{1}Nanjing University \quad
  \textsuperscript{2}SenseTime Research \quad
  \textsuperscript{3}ETH Zürich
}

\begin{document}

\maketitle

\begin{abstract}
Video matting is crucial for applications such as film production and virtual reality, yet deploying its computationally intensive models on resource-constrained devices presents challenges.
Quantization is a key technique for model compression and acceleration. As an efficient approach, Post-Training Quantization (PTQ) is still in its nascent stages for video matting, facing significant hurdles in maintaining accuracy and temporal coherence.
To address these challenges, this paper proposes a novel and general PTQ framework specifically designed for video matting models, marking, to the best of our knowledge, the first systematic attempt in this domain.
Our contributions include:
(1) A two-stage PTQ strategy that combines block-reconstruction-based optimization for fast, stable initial quantization and local dependency capture, followed by a global calibration of quantization parameters to minimize accuracy loss.
(2) A Statistically-Driven Global Affine Calibration (GAC) method that enables the network to compensate for cumulative statistical distortions arising from factors such as neglected BN layer effects, even reducing the error of existing PTQ methods on video matting tasks up to 20\%.
(3) An Optical Flow Assistance (OFA) component that leverages temporal and semantic priors from frames to guide the PTQ process, enhancing the model's ability to distinguish moving foregrounds in complex scenes and ultimately achieving near full-precision performance even under ultra-low-bit quantization.
Comprehensive quantitative and visual results show that our PTQ4VM achieves the state-of-the-art accuracy performance across different bit-widths compared to the existing quantization methods. We highlight that the 4-bit PTQ4VM even achieves performance close to the full-precision counterpart while enjoying 8$\times$ FLOP savings.
\end{abstract}

% \section{}

% Please read the instructions below carefully and follow them faithfully.

\begin{figure*}[htbp] % 使用 figure* 环境使其跨双栏（如果您的论文是双栏）
                      % 如果是单栏，可以使用 figure
    \centering % 使整个figure环境在其列中居中

    % 插入拼接后的单张图片
    \includegraphics[width=1\textwidth]{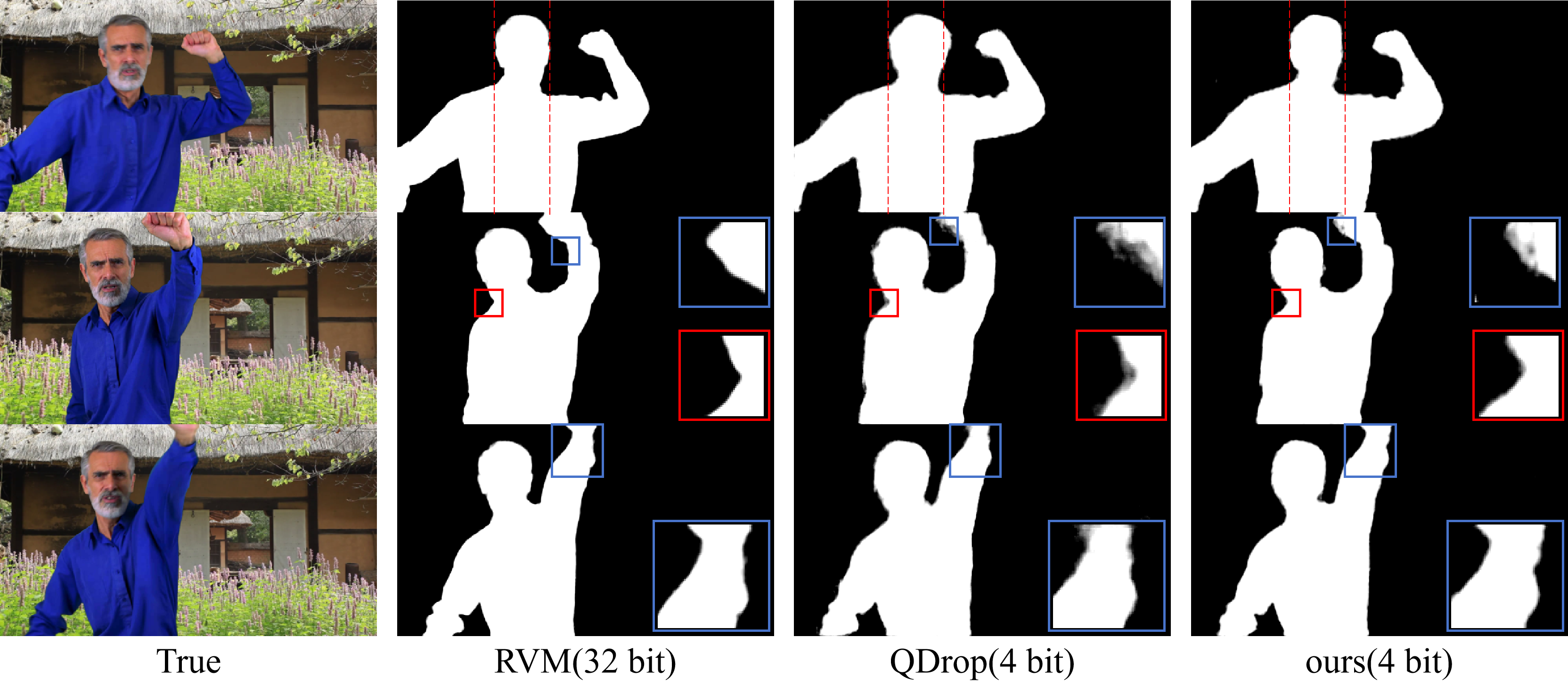} 
    \caption{Visual comparison of our PTQ4VM against Full-Precision (RVM) and QDrop . Our method demonstrates significant advantages in preserving fine details and temporal coherence.}
    \label{fig:intro_visual_comparison} % 整个figure的标签
\end{figure*}
\section{Introduction}

The purpose of video matting \cite{aksoy2017designing,bai2007geodesic,chen2013knn,chuang2001bayesian,feng2016cluster,li2024vmformer,lin2021real,lin2022robust,sengupta2020background,sun2021deep,zhang2021attention,zhao2021efficient,zhao2022discrete,zhao2023ddfm,zhao2023deep,zhao2023spherical,yao2024vitmatte} is to accurately estimate the alpha matte ($\alpha \in [0, 1]$) of the foreground objects for each frame in a video sequence. The alpha matte defines the foreground opacity at each pixel, governed by the compositing equation $I = \alpha F + (1-\alpha)B$, where $I$ is the observed pixel, $F$ is the foreground, and $B$ is the background. This challenging computer vision task has broad applications in film production, video conferencing, virtual reality, and more. To enable real-time performance and deployment on resource-constrained platforms for these diverse applications, efficient model representations are crucial. This necessitates advanced model compression techniques to reduce the computational and memory footprint of video matting models, making them suitable for edge computing devices.

% 中文注释部分:
% 视频抠图 (Video Matting) 旨在从视频序列的每一帧中精准地估计出前景对象的alpha遮罩 ($\alpha \in [0, 1]$)。
% 该alpha遮罩描述了每个像素属于前景的程度，其核心在于解决混合方程 $I = \alpha F + (1-\alpha)B$，其中 $I$ 是观测像素，$F$ 是前景，$B$ 是背景。
% 这是一项极具挑战性的计算机视觉任务，在电影制作、视频会议、虚拟现实等等领域有着广泛的应用前景。
% 为了在这些多样化的应用中实现实时性能并部署到资源受限的平台上，高效的模型表示至关重要。
% 这就需要先进的模型压缩技术来减少视频抠图模型的计算和内存占用，使其适用于边缘计算设备。

Model compression techniques, particularly quantization\cite{jacob2018quantization,nagel2021white,gholami2022survey}, are paramount for deploying advanced video matting models on resource-constrained devices by converting high-precision floating-point numbers to low-bit integers, thereby reducing model size and accelerating computation. While Quantization-Aware Training \cite{qin2023bimatting}(QAT) simulates quantization during training to achieve good performance, it demands extensive labeled data and computational resources, which are often scarce for video matting. In contrast, Post-Training Quantization (PTQ) quantizes pre-trained models directly with minimal calibration data and no retraining, offering significant advantages in deployment efficiency. However, dedicated Post-Training Quantization research for video matting models remains nascent. In this work, we aim to systematically investigate the challenges and opportunities of applying PTQ to video matting tasks.

% 中文注释:
% 模型压缩技术，特别是量化，通过将高精度浮点数转换为低比特整数，从而减少模型体积并加速计算，对于在资源受限设备上部署先进的视频抠图模型至关重要。
% 尽管量化感知训练 (QAT) 在训练过程中模拟量化效应以取得良好性能，但它需要大量的标注数据和计算资源，这对于视频抠图任务而言往往是稀缺的。
% 相比之下，训练后量化 (PTQ) 直接对预训练模型进行量化，仅需少量校准数据且无需重新训练，在部署效率上具有显著优势。
% 然而，针对视频抠图模型的专门训练后量化研究尚处于起步阶段。在本工作中，我们旨在系统性地研究将PTQ应用于视频抠图任务时所面临的挑战与机遇，重点关注保持单帧准确性和时序连贯性。
However, applying PTQ to complex video matting models presents challenges. Firstly, their deep topological structures and the reliance on limited calibration data often lead to unstable convergence during the PTQ calibration process. Secondly, at low bit-widths, quantization errors propagate through the network, resulting in artifacts and increased uncertainty in the output. Furthermore, recurrent structures, crucial for capturing temporal dependencies, are particularly vulnerable to quantization noise, which can destabilize learned temporal dynamics and manifest as flickering or jitter.
%然而，将PTQ应用于复杂的视频抠图模型，会面临多重难题。
%首先，其深层拓扑结构以及对少量校准数据的依赖，常常导致PTQ校准过程收敛不稳定。
%其次，在低比特下，量化误差会在网络中传播，导致明显的伪影和输出不确定性的增加。
%此外，对于捕捉时间依赖至关重要的循环结构，尤其容易受到量化噪声的影响，这会破坏学习到的时间动态的稳定性，并表现为闪烁或抖动。

% 为应对这些挑战，本文提出了一种专为视频抠图模型设计的创新PTQ框架。
% 据我们所知，这是首个系统性解决该任务PTQ问题的工作。
% 我们的框架旨在具有通用性，其主要贡献如下
To address these challenges, this paper proposes a novel PTQ framework specifically designed for video matting models. To the best of our knowledge, this is the first work to systematically tackle PTQ for this task. Our framework is designed to be general, and its main contributions are as follows:

\begin{enumerate}
    % 1.  结合块结构与全局优化的两阶段PTQ策略：我们首先采用基于块重构的优化方法对网络进行初步量化，该方法能够实现快速稳定的收敛并捕捉关键的局部依赖关系；随后，在全局视角下对量化参数进行进一步校准，从而在保持PTQ效率的同时，最大限度地减少精度损失。
    \item \textbf{A Two-Stage PTQ Strategy Combining Block-wise and Global Optimization} We initially quantize the network using block-wise optimization, which achieves fast and stable convergence while capturing critical local dependencies. Subsequently, we perform a global calibration of quantization parameters to minimize accuracy loss while preserving PTQ efficiency.

    % 2.  % 统计驱动的量化参数全局仿射校准：我们观察到，标准的训练后量化流程中对BN层的忽视往往会导致网络中间层输出分布发生显著的统计特性改变。我们提出了一种全局仿射校准 (Global Affine Calibration, GAC) 方法，使得网络能够学习补偿累积的统计失真。在开销极小的情况下显著提升了整体模型性能。
    \item \textbf{Statistically-Driven Global Affine Calibration of Quantization Parameters} We observe that neglecting Batch Normalization (BN) layers\cite{ioffe2015batch} in standard Post-Training Quantization (PTQ) pipelines often leads to significant statistical alterations in the distributions of intermediate layer outputs.  We propose a Global Affine Calibration (GAC) method that enables the network to learn a compensation for these cumulative statistical distortions. 

    % OFA: 利用光流辅助指导训练后量化：为了契合视频的时序与语义特性，我们创新性地引入了光流辅助 (Optical Flow Assistance, OFA) 组件。该组件利用从相邻帧计算得到的光流场，将前一帧的预测结果扭曲 (warp) 作为当前帧的强时序与语义先验。在该组件的引导下，PTQ过程增强了模型对复杂场景中运动前景与背景的区分能力，从而改善了整体的语义理解和抠图精度。
    \item \textbf{Optical Flow Assistance to Guide Post-Training Quantization} To align with the temporal and semantic characteristics of video, we innovatively introduce an Optical Flow\cite{horn1981determining} Assistance (OFA) component. This component utilizes optical flow fields computed from adjacent frames to warp the prediction of the previous frame, serving as a strong temporal and semantic prior for the current frame. Under the guidance of this component, the Post-Training Quantization (PTQ) process enhances the model's ability to distinguish between moving foregrounds and backgrounds in complex scenes. 
\end{enumerate}
Our proposed framework (PTQ4VM) not only quantitatively reduces the error of existing PTQ methods on video matting tasks by 10\%--20\% but also achieves performance remarkably close to the full-precision counterpart, even under challenging 4-bit quantization, while concurrently enjoying substantial $8\times$ FLOP savings.
\section{Related Work}
% 本节回顾了与我们工作最相关的三个领域：视频抠图技术、神经网络的训练后量化（PTQ）方法，以及用于增强时间一致性的光流技术。
% 深度学习的兴起极大地推动了视频抠图技术的发展，使其在精度和鲁棒性上超越了依赖低级手工特征的传统方法。
\textbf{Video Matting} is significantly developed with the advent of deep learning, surpassing traditional methods\cite{smith1996blue,chuang2002video} reliant on low-level handcrafted priors in terms of accuracy and robustness.
% 早期的深度学习模型开始探索利用时序信息，而一些精细的图像抠图工作也为视频领域提供了借鉴。
% Early deep learning models began exploring temporal information utilization\cite{chuang2002video}, while sophisticated image matting works also provided insights for the video domain.
% 现代视频抠图模型通常采用编码器-解码器架构，并结合特定模块来处理时间依赖性。例如一些模型专注于实时图像抠图，展示了单帧任务在效率上的进展。
Modern video matting models often employ encoder-decoder architectures augmented with specific modules to handle temporal dependencies. For instance, some models focus on real-time image matting, showcasing progress in efficiency for single-frame tasks.
% 在众多视频抠图模型中，RVM (Robust High-Resolution Video Matting with Temporal Guidance) 通过其循环结构（如GRU）和时间引导机制，有效地利用历史帧信息，实现了高分辨率视频的平滑抠图。
Among various video matting models, RVM (Robust High-Resolution Video Matting with Temporal Guidance)\cite{lin2022robust} stands out for its effective use of temporal information via a recurrent architecture (e.g., GRU) and temporal guidance mechanisms, achieving smooth mattes for high-resolution videos.
% RVM在精度、时间一致性和模型效率之间取得了良好平衡。相较于一些更庞大的模型，RVM在保持高质量输出的同时，其架构相对更为轻量。
RVM strikes a good balance between accuracy, temporal consistency, and model efficiency. Compared to some larger counterparts, RVM maintains high-quality output with a relatively lightweight architecture.
% 因此，我们选择RVM作为本文量化研究的全精度基准模型，因其是实现高效实际部署的理想候选者。
Consequently, we select RVM as the full-precision baseline model for our quantization study, as it is an ideal candidate for efficient practical deployment post-quantization.
% 然而，即便是如RVM这样相对高效的模型，在部署于资源受限设备时，其固有的计算和存储需求仍构成挑战，凸显了模型压缩（如量化）的必要性。
However, even relatively efficient models like RVM present computational and memory challenges when deployed on resource-constrained devices, underscoring the need for model compression techniques such as quantization.

\textbf{Post-Training Quantization (PTQ)} focuses on the accurate determination of the quantization parameters. MSE-based methods are foundational, optimizing $s$ and $z$ by minimizing the Mean Squared Error between the original floating-point tensors and their quantized counterparts using a calibration set.
% 为进一步提升PTQ性能，多种先进算法被提出。AdaRound学习权重值的最优舍入策略，使其向最小化任务损失而非仅权重重构误差的方向调整，对低比特量化尤为有效。
To further enhance PTQ performance, several advanced algorithms have been proposed. AdaRound\cite{nagel2020up} learns an optimal rounding strategy for weight quantization, adapting weights towards minimizing task loss rather than mere weight reconstruction error, proving particularly effective for very low bit-widths.
% BRECQ (Block Reconstruction) 将网络划分为块，并逐块优化量化参数以最小化块输出的重构误差，从而更好地捕获层间依赖性，优于传统的逐层量化。
BRECQ (Block Reconstruction)\cite{li2021brecq} improves upon layer-wise quantization by partitioning the network into blocks and optimizing quantization parameters per block to minimize the reconstruction error of its output, thereby better capturing inter-layer dependencies. 
% QDrop 则通过在校准时模拟量化噪声（例如，随机丢弃部分激活值的量化版本），增强模型对量化扰动的鲁棒性。
QDrop\cite{wei2022qdrop} enhances model robustness to quantization perturbations by simulating quantization noise during calibration, for instance, by randomly dropping quantized versions of activations.
% 尽管这些PTQ技术在通用视觉任务中表现出色，但将它们有效地组合并针对视频抠图的独特需求（如结合块优化与全局校准、特定权重调整策略以及时间一致性保持）进行调整和创新，仍是有待探索的领域。本文工作针对这些方面，提出了一个定制化的PTQ流程。
While these PTQ techniques demonstrate strong performance on general vision tasks, their optimal combination and adaptation for the unique demands of video matting, such as integrating block-wise optimization with global calibration, specific weight adjustment strategies, and preserving temporal consistency, remain open research areas. Our work addresses these aspects by proposing a tailored PTQ pipeline.

% 光流估计计算视频帧间的像素级运动，广泛应用于运动分析、物体跟踪、视频稳定及作为复杂视频理解任务（如视频抠图）的输入。
\textbf{Optical Flow} estimation computes pixel-level motion between video frames and is widely applied in motion analysis, object tracking, video stabilization, and as input for complex video understanding tasks such as video matting.
% 传统方法如Lucas-Kanade依赖局部约束。深度学习方法，自FlowNet起，通过CNN端到端学习光流，显著提升了精度与鲁棒性。后续如PWC-Net引入了特征金字塔和代价体积。
Traditional methods like Lucas-Kanade\cite{lucas1981iterative} rely on local constraints. Deep learning approaches, since FlowNet\cite{dosovitskiy2015flownet}, learn optical flow end-to-end via CNNs, significantly improving accuracy and robustness. Subsequent methods, such as PWC-Net\cite{sun2018pwc}, introduced feature pyramids and cost volumes.
% 在当前先进算法中，RAFT (Recurrent All-Pairs Field Transforms) 表现突出。
Among current state-of-the-art algorithms, RAFT (Recurrent All-Pairs Field Transforms)\cite{teed2020raft} exhibits outstanding performance.
% RAFT的核心在于其迭代优化机制：它构建所有像素对相关性的4D代价体积金字塔，并通过循环单元（如ConvGRU）从初始流场迭代更新光流估计。
The core of RAFT lies in its iterative optimization mechanism: it constructs a 4D cost volume pyramid of all-pairs correlations and iteratively updates the flow field from an initial estimate using a recurrent unit (e.g., ConvGRU).
% RAFT的主要优势包括能有效处理大位移运动、良好的泛化能力以及在标准基准上的高精度。
Key advantages of RAFT include its effectiveness in handling large displacements, strong generalization capabilities, and high accuracy on standard benchmarks.
% 其迭代特性也允许在精度和效率间权衡。
Its iterative nature also allows for a trade-off between accuracy and efficiency.
% 因此，我们选用RAFT来获取高精度的光流先验，以辅助视频抠图的时序与语义增强。
Consequently, we select RAFT to obtain high-precision optical flow priors to assist in the temporal and semantic enhancement of video matting.

% \section{Methods}
\begin{figure}[htbp] % 假设的figure环境
    \centering
    \includegraphics[width=\textwidth]{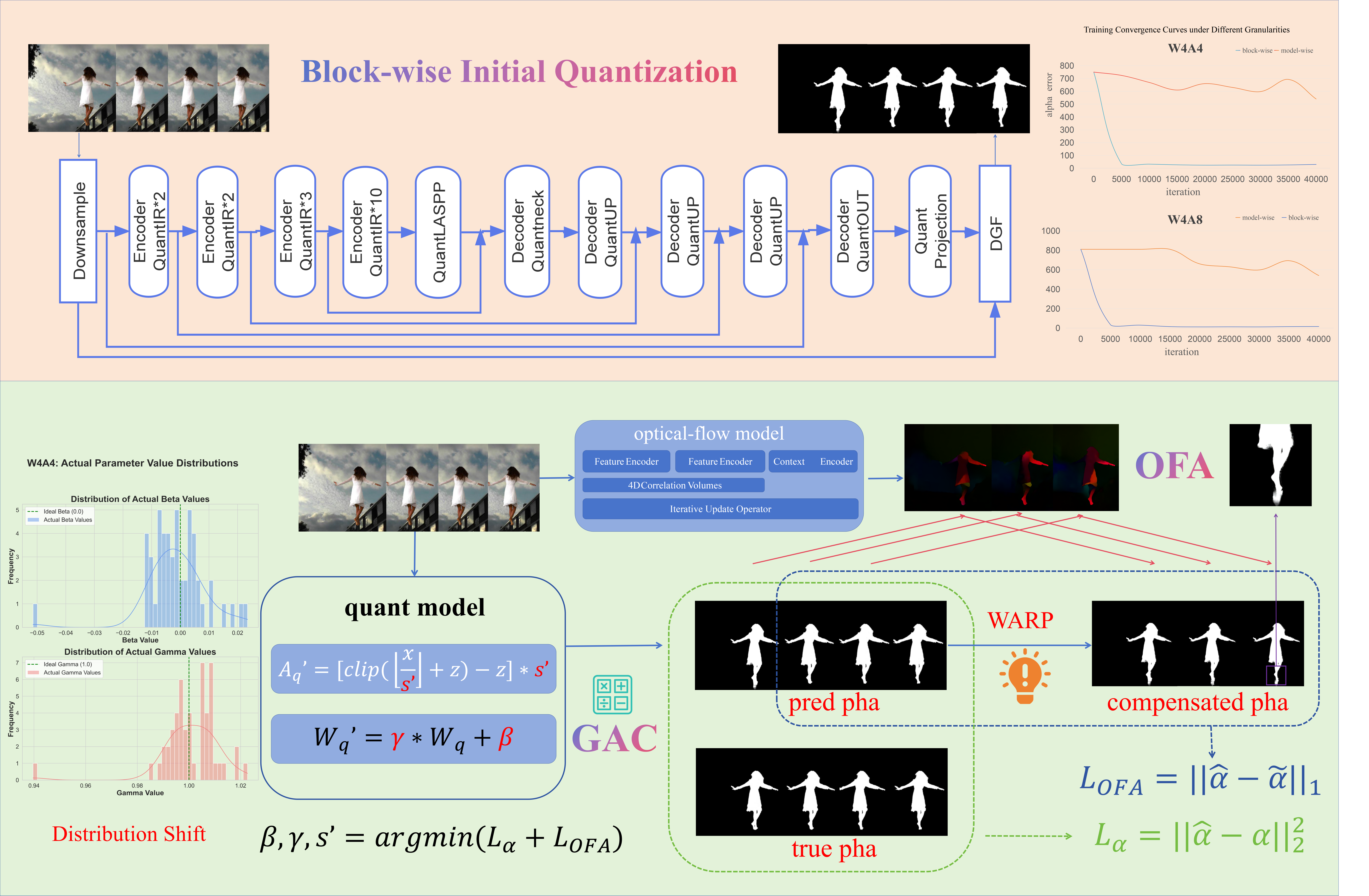} % 您的图片文件
    \caption{\footnotesize % 图片标题开始
        Overview of our PTQ4VM. % 我们PTQ4VM概览。
        In Stage 1 (Block-wise Initial Quantization), we optimize functional blocks sequentially to accelerate convergence and enhance stability. % 阶段一（逐块初始量化）：我们顺序优化功能块以加速收敛并增强初始量化的稳定性。
        In Stage 2 (Cross-Block Joint Calibration), we employ Global Affine Calibration (GAC) to compensate for distributional shifts; meanwhile, Optical Flow Assistance (OFA) guides the model to learn temporal-semantic coherence. % 阶段二（跨块联合校准）：我们首先采用全局仿射校准（GAC）来补偿分布偏移；同时，光流辅助（OFA）引导模型学习并增强最终视频抠图输出的时序语义连贯性。
    }
    \label{fig:overview_ptq4vm} % 图片的标签
\end{figure}

\section{Methods}

\subsection{Preliminaries}

\paragraph{Weight and Activation Quantization} % 中文注释: 权重量化与激活量化
% 权重量化的核心优化目标是最小化原始权重 $w_{fp}$ 与反量化后权重 $\hat{w}_{fp}$ 之间的差异
The fundamental principle of uniform affine quantization maps a full-precision value $v$ (e.g., FP32) to a lower-bit integer $v_q$ (e.g., INT8) using a scale factor $s$ and a zero-point $z$:
\begin{equation} \label{eq:ptq_quant_related_work_v2}
v_q = \text{clip}(\text{round}(v/s + z), Q_{\text{min}}, Q_{\text{max}})
\end{equation}
% 其中 $\text{round}(\cdot)$ 是舍入函数（例如，四舍五入到最近整数），$\text{clamp}(\cdot, Q_{\text{min}}, Q_{\text{max}})$ 将结果限制在目标整数范围内（例如，对于有符号INT8为 $[-128, 127]$）。对应的反量化操作则重构原始值的近似值：$v \approx s(v_q - z)$。PTQ的核心挑战在于，在没有重训练和仅有少量数据的情况下，为权重和激活值找到最优的 $s$ 和 $z$。
where $\text{round}(\cdot)$ is a rounding function (e.g., round-to-nearest), and $\text{clip}(\cdot, Q_{\text{min}}, Q_{\text{max}})$ constrains the result to the target integer range (e.g., $[-128, 127]$ for signed INT8). The corresponding dequantization reconstructs an approximation of the original value: $v \approx s(v_q - z)$. The core challenge in PTQ is to find optimal $s$ and $z$ for weights and activations with minimal data and no retraining.

The core optimization objective of weight quantization is to minimize the difference between the original weights $W_{fp}$ and the quantized weights ${W}_{q}$.
% 激活量化发生在网络中激活函数的输出之后，将浮点激活值 $a_{fp}$ 转换为低位宽整数 $a_q$。
Activation Quantization occurs after the output of activation functions in the network, converting floating-point activations $A_{fp}$ to low-bitwidth integers $A_q$.
% 此过程通常使用一个小的代表性校准数据集来收集激活值的统计信息（如范围），并据此确定最优的量化参数（如尺度因子 $s_a$），目标是使量化后的网络输出尽可能接近全精度网络的输出。
This process typically uses a small, representative calibration dataset to collect statistical information about activations (such as their range) and thereby determine optimal quantization parameters (like the scale factor $s$). The goal is to make the output of the quantized network as close as possible to that of the full-precision network.
\paragraph{Batch Normalization Folding} % 中文注释: 批归一化层折叠
% 在推理阶段，批归一化 (BN) 层的操作是线性的，可以与其前置的卷积层（或全连接层）的参数进行数学等价的融合，以减少计算量。
During inference, the operations of a Batch Normalization (BN) layer are linear and can be mathematically equivalent to being fused with the parameters of its preceding convolutional (or fully connected) layer to reduce computation.
% 设原始卷积层（或全连接层）的输出为 $Y = W X + B$ (其中 $W$ 为权重，$B$ 为偏置，$X$ 为输入)。
Let the output of the original convolutional (or fully connected) layer be $Y = WX + B$ (where $W$ are weights, $B$ is bias, and $X$ is input).
% 其后的BN层操作（使用推理时的固定参数）为 $Y_{\text{BN}} = \gamma \frac{Y - \mu}{\sqrt{\sigma^2 + \epsilon}} + \beta$，其中 $\mu$ 和 $\sigma^2$ 是BN层累积的均值和方差，$\gamma$ 和 $\beta$ 是可学习的缩放和平移参数，$\epsilon$ 是一个防止除零的小常数。
The subsequent BN layer operation (using fixed parameters at inference) is $Y_{\text{BN}} = \gamma \frac{Y - \mu}{\sqrt{\sigma^2 + \epsilon}} + \beta$, where $\mu$ and $\sigma^2$ are the accumulated mean and variance of the BN layer, $\gamma$ and $\beta$ are learnable scale and shift parameters, and $\epsilon$ is a small constant to prevent division by zero.

% 通过折叠，可以得到新的等效权重 $w_f$ 和偏置 $B_f$：
Through folding, new equivalent weights $w_f$ and bias $B_f$ can be obtained:$W_f = \frac{\gamma W}{\sqrt{\sigma^2 + \epsilon}}$, $B_f = \frac{\gamma (B - \mu)}{\sqrt{\sigma^2 + \epsilon}} + \beta$
% \begin{align}
%     W_f &= \frac{\gamma W}{\sqrt{\sigma^2 + \epsilon}} \label{eq:bn_fold_w} \\
%     B_f &= \frac{\gamma (B - \mu)}{\sqrt{\sigma^2 + \epsilon}} + \beta \label{eq:bn_fold_b}
% \end{align}
% 使得折叠后的层输出 $Y' = w_f X + B_f$ 在数学上等价于 $Y_{\text{BN}}$。
such that the output of the folded layer $Y' = W_f X + B_f$ is mathematically equivalent to $Y_{\text{BN}}$.
% 在全精度模型中，此BN折叠是无损的。
In full-precision models, this BN folding is lossless.
% 但是在量化时，有关BN层的修正往往被人们忽略，我们将在3.3节详细讨论。
However, during quantization, corrections related to the BN layer are often overlooked. We will discuss this in detail in Section~3.3.

% \begin{figure}[t] % 假设的figure环境
%     \centering
%     \includegraphics[width=\textwidth]{01_4.png} % 您的图片文件
%     \caption{\footnotesize % 图片标题开始
%         Overview of our PTQ4VM. % 我们PTQ4VM概览。
%         In Stage 1 (Block-wise Initial Quantization), we optimize functional blocks sequentially to accelerate convergence and enhance stability. % 阶段一（逐块初始量化）：我们顺序优化功能块以加速收敛并增强初始量化的稳定性。
%         In Stage 2 (Cross-Block Joint Calibration), we employ Global Affine Calibration (GAC) to compensate for distributional shifts; meanwhile, Optical Flow Assistance (OFA) guides the model to learn temporal-semantic coherence. % 阶段二（跨块联合校准）：我们首先采用全局仿射校准（GAC）来补偿分布偏移；同时，光流辅助（OFA）引导模型学习并增强最终视频抠图输出的时序语义连贯性。
%     }
%     \label{fig:overview_ptq4vm} % 图片的标签
% \end{figure}

\subsection{BIQ: Block-wise Initial Quantization for Fast Convergence \& Local Dependency}

% **2. 优化粒度考量**
\paragraph{Consideration of Optimization Granularity}
The choice of optimization granularity is a critical factor affecting final quantization performance.
% 在对复杂模型结构进行PTQ量化时，量化噪声会有显著的影响。
When applying PTQ to models with complex architectures, quantization noise can have a significant impact.
% Efficient models with depth-wise separable convolutions show a significant drop for PTQ or even result in random performance.
Some studies\cite{nagel2019data} indicate that efficient models, particularly those with depth-wise separable convolutions, often exhibit a significant performance drop with PTQ, sometimes even resulting in random-level performance. 
% 实验也证明，直接进行优化将会面临训练不稳定、难收敛的情况。
Our experiments also confirm that attempting direct end-to-end optimization faces challenges such as training instability and convergence difficulties.
% 同时，layer-wise的校准忽略了层间的依赖关系，而且在视频任务中会有很高的显存要求。
Concurrently, layer-wise calibration overlooks inter-layer dependencies and can impose high memory requirements, especially in video tasks.
% 我们最终选择了块级的划分。
We ultimately opted for a block-wise partitioning strategy.
% 实验表明，块级优化不仅在计算效率上表现出色，能有效捕获关键的局部依赖关系，而且通过适当的块划分，能够保持最高的优化潜力，在性能和效率之间取得了最佳平衡。
Experiments indicate that block-wise optimization not only excels in computational efficiency and effectively captures crucial local dependencies but also, with appropriate block partitioning, maintains high optimization potential, striking an optimal balance between accuracy and efficiency.

% **1. 逐块顺序优化与参数学习**
\paragraph{Block-wise Sequential Optimization and Parameter Learning}
% 我们将网络划分为若干计算块 ($B_i$)元，并按拓扑顺序逐块量化。
We partition the network into several computational blocks ($B_i$), and quantize them sequentially in topological order.
% 对当前块 $B_i$：其量化版本的输入 $x_{q,in}$ 来自先前已量化块的输出；其全精度版本的对应输入 $x_{fp,in}$ 来自全精度前序块的输出（两者均源自同一原始校准样本）。
For the current block $B_i$, the input to its quantized version, $x_{q,in}$, is the output from preceding quantized blocks. Its full-precision counterpart receives a corresponding input, $x_{fp,in}$, from preceding full-precision blocks; both $x_{q,in}$ and $x_{fp,in}$ originate from the same raw calibration sample.

% 对于每个块 $B_i$，我们为其全精度权重 $W$ 学习最优的舍入参数，并为其输入激活学习自适应的尺度因子。这些量化参数通过迭代优化来确定，目标是最小化该块的量化输出 $Y_q$ 与其全精度输出 $Y_{fp}$ 之间的均方误差(MSE)。具体而言，$Y_{fp}$ 是使用输入 $x_{fp,in}$ 和原始权重 $W$ 计算得到的，而 $Y_q$ 则是使用输入 $x_{q,in}$ 以及学习到的权重和激活的量化表示计算得到的。目标函数为：
For each block $B_i$, we learn optimal rounding for its full-precision weights $W$ and learn adaptive scale factors for its input activations. These quantization parameters are determined by iteratively minimizing the Mean Squared Error (MSE) between the block's quantized output $Y_q$ and its full-precision output $Y_{fp}$. This learning process is performed iteratively over the calibration data.

% % % % % % % % % % % % % % % % % % % % % % % % % % % % % % % % % % % % % % % % % % % % % % % % % % % % % % % % % % % % %
% subsection 3.x 权重量化的线性校准
% % % % % % % % % % % % % % % % % % % % % % % % % % % % % % % % % % % % % % % % % % % % % % % % % % % % % % % % % % % % %
\subsection{GAC: Global Affine Calibration for Statistical Deviations in PTQ}
% % % % % % % % % % % % % % % % % % % % % % % % % % % % % % % % % % % % % % % % % % % % % % % % % % % % % % % % % % % % %
% paragraph 误差分析：量化后网络中间输出的分布偏移
% % % % % % % % % % % % % % % % % % % % % % % % % % % % % % % % % % % % % % % % % % % % % % % % % % % % % % % % % % % % %
% #### 误差分析：量化后网络中间输出的分布偏移
\paragraph{Distributional Shift of Intermediate Outputs post-Quantization}

% 量化后的精度严重下降是PTQ的核心问题。我们首次将视角聚焦于BN层，从统计分析的角度解释了这一现象，并指出通用PTQ框架的不足。
The core issue in PTQ is the significant accuracy degradation post-quantization. We are the first to focus on the Batch Normalization (BN) layer, explaining this phenomenon from a statistical analysis perspective and highlighting the shortcomings of general PTQ frameworks.

% 通用的后训练量化 (Post-Training Quantization, PTQ) 框架通常首先将批量归一化 (Batch Normalization, BN) 层折叠（融合）到其前驱的卷积层或全连接层中，形成有效权重 $w_f$，随后再对 $w_f$ 进行权重量化。
Typical Post-Training Quantization (PTQ) frameworks initially fold Batch Normalization (BN) layers into their preceding convolutional or fully-connected layers, yielding effective weights $W_f$. Subsequently, these effective weights $W_f$ undergo weight quantization.

% 然而，我们观察到，在逐层量化并进行前向传播的过程中，由于权重和激活值量化引入的误差会逐层累积，导致网络中间层激活（即下一层的输入）$x$ 的统计特征（如均值、方差、分布形态）会发生明显偏移，偏离其在全精度网络中的对应状态。
However, we observe that during layer-wise quantization and forward propagation, the errors introduced by weight and activation quantization accumulate. This accumulation leads to a significant shift in the statistical characteristics (e.g., mean, variance, distribution shape) of intermediate layer activations $x$ (i.e., the input to the next layer), causing them to deviate from their counterparts in the full-precision network.
% 当这种发生偏移的激活 $x$ 与根据原始全精度统计信息折叠得到的权重 $w_f$ 进行运算时，$w_f$ 对于当前真实的输入分布而言已不再是最优的。
When these shifted activations $x$ are processed with the folded weights $W_f$ (which were derived based on original full-precision statistics), $W_f$ is no longer optimal for the actual input distribution it encounters.
% 相应地，以最小化原始 $w_f$ 与其量化版本 $\hat{w}_f$ 之间差异为目标的传统权重量化策略，也因未考虑输入分布的偏移而变得次优。
Consequently, conventional weight quantization strategies aiming to minimize the difference between the original $W_f$ and its quantized version $\hat{W}_f$ become suboptimal, as they fail to account for this input distribution shift.

% 更为关键的是，这种累积的分布失真在经过非线性激活函数（例如ReLU、tanh等）时，其统计特性会被进一步重塑和改变。
Critically, such accumulated distributional distortion is further reshaped and altered when passed through non-linear activation functions (e.g., ReLU, Tanh).
% 这使得后续的激活量化步骤面临严峻挑战，因为激活量化器通常采用均匀量化，依赖于输入激活值的简单统计量（如观测到的最小值和最大值）来确定量化范围和尺度。
This poses a significant challenge for subsequent activation quantization steps, as activation quantizers typically employ uniform quantization, relying on simple statistics of the activations, such as observed minimum and maximum values, to determine quantization ranges and scales.
% 当输入激活的分布已经严重偏离“常规”或预期分布时，这些基于简单统计的量化器难以有效补偿这种失真，从而可能导致显著的精度下降。
When the activation distribution has substantially deviated from its "canonical" or expected form, these statistically driven quantizers struggle to effectively compensate for distortions, potentially leading to considerable accuracy degradation.

% % % % % % % % % % % % % % % % % % % % % % % % % % % % % % % % % % % % % % % % % % % % % % % % % % % % % % % % % % % % %
% paragraph 方法介绍：权重的全局仿射校准
% % % % % % % % % % % % % % % % % % % % % % % % % % % % % % % % % % % % % % % % % % % % % % % % % % % % % % % % % % % % %
\paragraph{Global Affine Calibration of Dequantized Weights}

Some works\cite{nagel2019data} have noted the bias in the quantization process and proposed pre-training Cross-Layer Equalization and Absorbing high biases.However, in our experiments, these methods did not yield any performance improvements when applied to the relatively complex models under our investigation. We attribute this primarily to the fact that in complex model architectures, quantization errors propagate layer by layer and are reshaped and amplified by non-linear operations. Consequently, merely adjusting weights quantitatively before quantization struggles to achieve satisfactory results. Therefore, we propose a more general global linear calibration method that directly adjusts the quantized weights.

% 然而，在我们的实验中，上述方法在应用于我们研究的相对复杂的模型时，并未展现出任何效果提升。我们分析这主要原因在于：当模型结构较为复杂时，量化误差会在网络中逐层传播并被非线性运算重塑放大，此时仅在量化前对权重进行定量调整难以取得满意的结果。因此，我们提出了一种更为通用的全局线性校准方法，直接对量化后的权重进行调整。

% 我们的方法如下：对于网络中的**每一个卷积层** $i$，我们引入两个层级的标量校准参数用于权重：缩放因子 $\gamma_i$ 和平移因子 $\beta_i$。这两个参数将应用于该层对应的初始量化后的折叠权重 $W_{f,q,i}$：
Our method is as follows: for each convolutional layer $i$ in the network, we introduce two scalar calibration parameters for weights: a scaling factor $\gamma_i$ and a shift factor $\beta_i$. These parameters are applied to the corresponding initially quantized folded weights $W_{f,q,i}$ of that layer:
\begin{equation}
    W'_{f,q,i} = \gamma_i W_{f,q,i} + \beta_i
    \label{eq:affine_weight_calibration_final_revised_params}
\end{equation}
% 类似地，对于输入到第 $i$ 层的激活值 $x_i$，其应用量化函数后的表示可以表示为，其中 $s'_{a,i}$ 是我们优化的激活尺度因子，$z'_{a,i}$ 是预先确定的零点：
Similarly, for activations $x_i$ input to layer $i$, their representation after applying the quantization function, where $s'_{a,i}$ is the activation scaling factor we optimize and $z'_{a,i}$ is a pre-determined zero-point, can be expressed as:
\begin{equation}
    A'_{q,i} = ({\clip}(\lfloor x_i/s'_{a,i} \rfloor + z_{a,i}, Q_{\min,a}, Q_{\max,a}) - z_{a,i}) \cdot s'_{a,i}
    \label{eq:activation_quant_dequant}
\end{equation}
% 其中 $Q_{\min,a}$ 和 $Q_{\max,a}$ 是激活量化的裁剪边界。
where $Q_{\min,a}$ and $Q_{\max,a}$ are the clipping bounds for activation quantization.

% 权重校准参数 $\{\gamma_i\}$、$\{\beta_i\}$ 以及激活尺度因子 $\{s'_{a,i}\}$ (对于所有相关层) 通过联合优化得到，目标是最小化网络最终输出的预测alpha值 (记为 $\hat{\alpha}$) 与真实透明度 (记为 $\alpha$) 之间的均方误差 (MSE)。这个损失是在校准数据集 $D_{\text{calib}}$ 上计算的。
The weight calibration parameters $\{\gamma_i\}$, $\{\beta_i\}$, along with the activation scaling factors $\{s'_{a,i}\}$ for all relevant layers, are jointly optimized by minimizing the Mean Squared Error (MSE) between the network's final predicted alpha values ($\hat{\alpha}$) and the ground truth alpha mattes ($\alpha$). 

% 在校准完成后，这些学习到的参数 $\gamma_i$、$\beta_i$ 和 $s'_{a,i}$ 可以方便地融入对应层权重 $W_{f,q,i}$ 的量化参数以及激活的量化参数中，因此在推理阶段通常不会对模型引入新的参数量或显著的额外计算开销。
After calibration, these learned parameters $\gamma_i$, $\beta_i$, and $s'_{a,i}$ can be conveniently absorbed into the quantization parameters of the corresponding layer's weights $W_{f,q,i}$ and activations, respectively. Thus, they typically introduce no new parameters or significant additional computational overhead during inference.

% 此端到端优化过程使得各层的 $\gamma_i$、$\beta_i$ 和 $s'_{a,i}$ 能够协同学习一种全局性的补偿机制，系统性地修正由量化操作引入的累积误差和分布偏移。
This end-to-end optimization process enables the layer-specific $\gamma_i$, $\beta_i$, and $s'_{a,i}$ to collaboratively learn a global compensation mechanism, systematically correcting accumulated errors and distributional shifts introduced by quantization.
% 该方法具有良好的通用性，因为它不依赖于对特定层或特定类型误差的复杂建模，而是通过优化最终任务目标来直接调整整体权重尺度和偏置。
The method exhibits good universality as it does not rely on complex modeling of specific layers or error types but directly adjusts overall weight and activation scales and biases by optimizing the final task objective.
% 值得强调的是，我们的全局校准机制可以便捷地应用于各种现有的PTQ方法之上，并能在这些方法的基础上带来显著的性能提升。
Importantly, our global calibration mechanism can be readily applied on top of various existing PTQ methods, yielding significant performance improvements.

% % % % % % % % % % % % % % % % % % % % % % % % % % % % % % % % % % % % % % % % % % % % % % % % % % % % % % % % % % % % %
% subsection 3.y 基于光流的时序一致性优化 (Temporal Consistency Refinement using Optical Flow)
% % % % % % % % % % % % % % % % % % % % % % % % % % % % % % % % % % % % % % % % % % % % % % % % % % % % % % % % % % % % %
% ### 光流辅助的视频抠图时序与语义增强技术
\subsection{OFA: Optical Flow Assistance for Temporal-Semantic Coherence in PTQ}

% 在视频抠图任务中，尤其对于量化模型而言，仅仅逐帧独立预测 $\alpha$ matte 往往难以捕捉复杂的动态场景，容易导致输出结果在时间维度上出现闪烁或不连贯的瑕疵。
In video matting tasks, particularly for quantized models, merely predicting $\alpha$ mattes frame-by-frame often fails to capture complex dynamic scenes, leading to temporal flickering or inconsistencies in the output.
% 为进一步提升预测结果的质量，我们创新性的引入了基于光流的优化方法。
To further enhance the quality of predictions, we innovatively introduce an optimization method based on optical flow.
% 光流不仅能够提供强大的时序一致性约束，通过捕捉像素级的运动轨迹来平滑连续帧之间的过渡，还能辅助模型进行更深层次的语义识别与运动语义理解。
Optical flow not only provides robust temporal consistency constraints by capturing pixel-level motion trajectories to smooth transitions between consecutive frames, but also assists the model in deeper semantic recognition and motion semantic understanding.
% 值得注意的是，尽管光流估计本身具有一定的计算复杂度，这使其在需要大量迭代的训练场景（如全精度模型从头训练或量化感知训练 QAT）中不曾被直接集成，但后训练量化（PTQ）通常仅需一个非常小规模的校准数据集。

It is noteworthy that although optical flow estimation itself entails a certain computational complexity, which has precluded its direct integration into training scenarios requiring extensive iterations (such as training full-precision models from scratch or Quantization-Aware Training, QAT), Post-Training Quantization (PTQ) typically requires only a very small calibration dataset.
% 这种对数据量需求小、训练迭代周期短的特性，使得在PTQ框架内应用光流进行时序与语义增强变得计算上可行且目标明确。
This characteristic of low data demand and short training iteration cycles makes the application of optical flow for temporal and semantic enhancement computationally feasible and well-targeted within the PTQ framework.

% #### 方法原理
\paragraph{Method}
% 核心思想是利用视频帧间的运动信息来约束连续帧 $\alpha$ matte 的预测。
The core idea is to utilize inter-frame motion information to impose temporal constraints on $\alpha$ matte predictions across consecutive frames.
% 光流能够捕捉相邻输入图像帧 $I_{t-1}$ 和 $I_t$ 之间的像素级运动轨迹。
Optical flow captures pixel-level motion trajectories between adjacent input frames $I_{t-1}$ and $I_t$.
% 通过计算 $I_{t-1}$ 到 $I_t$ 的光流场 $F_{t-1 \rightarrow t}$，前一帧由模型预测的 $\alpha$ matte $\hat{\alpha}_{t-1}$ 可被有效地“扭曲” (warp) 到当前帧的坐标系下，形成对当前帧 $\alpha$ matte 的一个基于运动补偿的估计 $\tilde{\alpha}_t = \text{Warp}(\hat{\alpha}_{t-1}, F_{t-1 \rightarrow t})$。
By computing the optical flow field $F_{t-1 \rightarrow t}$ from $I_{t-1}$ to $I_t$, the $\alpha$ matte $\hat{\alpha}_{t-1}$ predicted by the model for the previous frame can be effectively warped to the coordinate system of the current frame, yielding a motion-compensated estimate for the current frame's $\alpha$ matte: $\tilde{\alpha}_t = \Warp(\hat{\alpha}_{t-1}, F_{t-1 \rightarrow t})$.

% 此经光流扭曲得到的 $\tilde{\alpha}_t$ 可视为对当前帧真实 $\alpha$ matte 的一个强时序先验。
This flow-warped matte, $\tilde{\alpha}_t$, serves as a strong temporal prior for the current frame's true $\alpha$ matte.
% 我们期望模型对当前帧 $I_t$ 直接预测的 $\hat{\alpha}_t = M_Q(I_t)$ (其中 $M_Q$ 是量化模型) 与该运动补偿先验 $\tilde{\alpha}_t$ 保持一致。
We encourage the model's direct prediction for the current frame, $\hat{\alpha}_t = M_Q(I_t)$ (where $M_Q$ is the quantized model), to align with this motion-compensated prior $\tilde{\alpha}_t$.
% 这种一致性通过 L1 损失函数量化，并作为正则项整合进模型的优化目标中，通常用于微调从阶段一获得的参数或在特定的PTQ优化阶段应用。
This alignment is quantified using an L1 loss, which is incorporated as a regularization term into the model's optimization objective, typically for fine-tuning parameters obtained from Phase 1 or during a dedicated PTQ optimization.% 通过预计算和存储小型校准集上的光流F，L_OFA的计算变得非常简洁、迅速。轻便的OFA组件进一步提升了我们ptq框架的优越与高效。
By pre-computing and storing the optical flow $F$ on the small calibration set, the computation of $\mathcal{L}_{\text{OFA}}$ becomes very concise and rapid. This lightweight OFA component further enhances the superiority and efficiency of our PTQ framework.

% #### 具体流程与损失函数
\paragraph{Procedure and Loss Function}
% 给定视频序列中的连续两帧图像 $I_{t-1}$ 和 $I_t$：
Given two consecutive frames $I_{t-1}$ and $I_t$ from a video sequence:
\begin{enumerate}
    % % 中文注释：1. 光流计算：采用RAFT算法计算从 $I_{t-1}$ 到 $I_t$ 的光流场 $F_{t-1 \rightarrow t}$。
    \item \textbf{Optical Flow Estimation:} Compute the optical flow field $F_{t-1 \rightarrow t}$ from $I_{t-1}$ to $I_t$ using the RAFT algorithm.
    % % 中文注释：2. 前帧 Alpha 预测：获取模型对前一帧 $I_{t-1}$ 预测的 alpha matte $\hat{\alpha}_{t-1} = M_Q(I_{t-1})$。
    \item \textbf{Previous Frame Alpha Prediction:} Obtain the model's predicted alpha matte for the previous frame, $\hat{\alpha}_{t-1} = M_Q(I_{t-1})$.
    % % 中文注释：3. Alpha Warping (扭曲)：使用步骤 1 中得到的光流场 $F_{t-1 \rightarrow t}$ 对 $\hat{\alpha}_{t-1}$ 进行扭曲操作，得到运动补偿后的 alpha matte：$\tilde{\alpha}_t = \text{Warp}(\hat{\alpha}_{t-1}, F_{t-1 \rightarrow t})$。
    \item \textbf{Alpha Warping:} Warp $\hat{\alpha}_{t-1}$ using the estimated flow field $F_{t-1 \rightarrow t}$ to obtain the motion-compensated alpha matte: $\tilde{\alpha}_t = \text{Warp}(\hat{\alpha}_{t-1}, F_{t-1 \rightarrow t})$.
    % % 中文注释：4. 当前帧 Alpha 预测：获取模型对当前帧 $I_t$ 直接预测的 alpha matte $\hat{\alpha}_t = M_Q(I_t)$。
    \item \textbf{Current Frame Alpha Prediction:} Obtain the model's direct prediction for the current frame, $\hat{\alpha}_t = M_Q(I_t)$.
    % % 中文注释：5. 光流辅助损失 (Optical Flow Assisted Loss)：计算 $\hat{\alpha}_t$ 与 $\tilde{\alpha}_t$ 之间的 L1 损失，作为光流辅助损失 $\mathcal{L}_{OFA}$：
    \item \textbf{Optical Flow Assisted Loss:} Calculate the L1 distance between $\hat{\alpha}_t$ and $\tilde{\alpha}_t$ to define the Optical Flow Assisted (OFA) loss: $\mathcal{L}_{\text{OFA}} = \| \hat{\alpha}_t - \tilde{\alpha}_t \|_1$
\end{enumerate}
% \begin{equation}
%     \mathcal{L}_{\text{OFA}} = \| \hat{\alpha}_t - \tilde{\alpha}_t \|_1
%     \label{eq:ofa_loss}
% \end{equation}
% % 中文注释：此损失项 $\mathcal{L}_{OFA}$ 被整合入网络的整体优化目标中，用以指导模型（或在量化参数微调阶段）生成时序上更连贯、语义上更准确的 alpha matte。
This loss term $\mathcal{L}_{\text{OFA}}$ is incorporated into the network's overall optimization objective to guide the model (or during a quantization parameter fine-tuning stage) towards generating more temporally coherent and semantically accurate alpha mattes.
% \begin{equation}
%     \mathcal{L} = \mathcal{L}_{\alpha} + \lambda \mathcal{L}_{\text{OFA}}
%     \label{eq:overall_loss}
% \end{equation}

\section{Experiments}
% \subsection{Setup}

% **Data Construction**
\paragraph{Data Construction}
% 我们的校准集非常小，仅从VM数据集中截取了256张图片（来自64个不同的视频片段，每个视频采样4帧）。
Our calibration set is very small, consisting of only 256 images sampled from the VM dataset\cite{lin2021real} (4 frames from each of 64 distinct video clips).
% 我们的评测集包括VM视频抠图数据集和D646图像抠图数据集。
Our evaluation datasets include the VM video matting dataset and the D646 image matting dataset\cite{qiao2020attention}.
% 值得注意的是，D646是一个模型在训练阶段未接触过的数据集，我们用它来进一步验证量化模型在抠图任务上的泛化能力与综合性能。
Notably, D646 is an image matting dataset unseen by the model during training, which we use to further validate the generalization ability and overall performance of the quantized model on matting tasks.

% **Evaluation Metrics**
\paragraph{Evaluation Metrics}
% 我们使用标准的视频/图像抠图评估指标来评估alpha matte ($\alpha$) 的质量，包括绝对差异总和 (Sum of Absolute Differences, SAD)、均方误差 (Mean Squared Error, MSE)、空间梯度误差 (spatial Gradient error, Grad) 和连通性误差 (Connectivity error, Conn)。
We assess the quality of the alpha matte ($\alpha$) using standard video/image matting evaluation metrics, including Sum of Absolute Differences (SAD), Mean Squared Error (MSE), spatial Gradient error (Grad), and Connectivity error (Conn).
% 对于视频数据集 (VM)，我们还使用时序平滑alpha差异偏差 (Deviation of Temporally Smoothed Alpha Differences, DTSSD) 来衡量时间相干性。
For the video matting (VM) dataset, we also measure temporal coherence using the Deviation of Temporally Smoothed Alpha Differences (DTSSD).

% **Implementation Details**
\paragraph{Implementation Details}
% 我们对模型中全部的卷积层和全连接层（若存在）进行量化。
We quantize all convolutional and fully-connected layers (if any) in the model.
% 在本研究中，主要探索的量化位宽范围是4至8比特 (4-8bit)。
In this study, the primary quantization bit-width range explored is 4 to 8 bits (4-8bit).
% 在第一阶段的逐块优化 (BIQ) 中，我们将原始RVM模型中的 InvertedResidual, LRASPP, BottleneckBlock, UpsamplingBlock, OutputBlock, Projection 等组件封装定义为独立的优化块。
During the first stage, Block-wise Initial Quantization (BIQ), we define components from the original RVM model, such as InvertedResidual, LRASPP, BottleneckBlock, UpsamplingBlock, OutputBlock, and Projection, as independent optimization blocks.% 在第二阶段中，我们对预训练好的量化模型进行全局的仿射微调，同时用OFA组件进行引导和约束。
During the second stage, we perform global affine fine-tuning on the pre-trained quantized model, concurrently guided and constrained by the OFA component.

% **Compared Methods**
\paragraph{Compared Methods}
% 为了验证我们方法的有效性，我们选取了当前领域内几种具有代表性的PTQ算法作为比较基线，包括：基于最小化均方误差的朴素量化 (MSE-based Quantization)、BRECQ 和 QDrop。
To validate the effectiveness of our method, we select several representative PTQ algorithms from the current field as comparison baselines, including: naive quantization based on minimizing Mean Squared Error (MSE-based Quantization), BRECQ\cite{li2021brecq}, and QDrop\cite{wei2022qdrop}(SOTA).
% 我们对每种对比方法都进行了广泛的参数调整，以确保它们达到各自理想的性能水平，从而保证比较的公平性。
We performed extensive parameter tuning for each comparison method to ensure they achieve their respective optimal performance levels, thereby guaranteeing a fair comparison.

\subsection{Main Results}
% 如表X所示，我们的PTQ方法在VM和D646两个数据集上的各项评估指标中均展现出显著优势。
As shown in Table~\ref {main}, our PTQ method demonstrates significant advantages across all evaluation metrics on both the VM and D646 datasets.
% 在8比特量化设置 (W8A8) 下，我们的方法能够取得与FP32全精度模型相媲美甚至在部分指标上更优的性能。
Under the 8-bit quantization setting (W8A8), our method achieves performance comparable to, and in some metrics even superior to, the FP32 full-precision model.
% 在更具挑战性的4比特量化场景，多种主流PTQ方法性能出现大幅度下降甚至趋于崩溃，而我们的方法依旧保持了令人满意的抠图质量和时序连贯性，显著优于其他对比方法。
In the more challenging 4-bit quantization scenario, where many mainstream PTQ methods exhibit substantial performance degradation or even collapse, our method still maintains satisfactory matting quality and temporal coherence, significantly outperforming other compared methods.
% 例如，在VM数据集的W4A4设置下，我们的方法在Alpha的各项误差指标上相比于次优方法有约20%的下降。
For instance, under the W4A4 setting on the VM dataset, our method shows a reduction of approximately 20\% in various alpha error metrics compared to the next best method.
% 这种在极低比特下的鲁棒性，体现了我们整体量化框架在处理复杂模型和误差累积方面的优越性。
This robustness at very low bit-widths highlights the superiority of our overall quantization framework in handling complex models and error accumulation.
% 尤为值得一提的是在D646数据集上的表现。由于我们的校准集完全来自VM视频数据集，D646对于模型而言是未校准的图像抠图数据。
Particularly noteworthy is the performance on the D646 dataset. Since our calibration set is derived entirely from the VM video dataset, D646 represents uncalibrated image matting data for the model.
% 我们的方法在该数据集上依旧取得了领先的量化性能，这充分证明了所提出方法的良好泛化能力，其核心校准策略能够有效地迁移到不同的数据分布和任务特性上。
Our method still achieves leading quantization performance on this dataset, which strongly demonstrates the good generalization ability of the proposed method, whose core calibration strategies can be effectively transferred to different data distributions and task characteristics.
% 总体而言，我们的方法在大幅压缩模型大小、降低计算复杂度的同时，最大限度地保留了视频抠图的精度和时序质量，为PTQ技术在复杂视频处理任务中的实际应用提供了有力的支持。
Overall, our method preserves the accuracy and temporal quality of video matting while substantially compressing model size and reducing computational complexity, providing robust support for the practical application of PTQ techniques in complex video processing tasks.
% 我们还提供了可视化的比较。如图~\ref{fig:visual_accuracy_comparison} (a)所示，我们的训练框架对抠图准确性的提升，在复杂的曲线细节上有更好的表现。如图~\ref{fig:visual_semantic_understanding} (b)所示，展示了我们的训练框架对视频语义理解的提升。对于相似的静态背景干扰，全精度模型有时也无法分辨，但是我们的模型很准确的识别了运动前景，这也印证了OFA组件的引导作用。

\begin{figure}[tbp] % 'h'尽量在此处, 't'顶部, 'b'底部, 'p'单独一页
    \centering % 使整个figure环境在其列中居中

    % 第一个子图 (a)
    \begin{subfigure}[b]{0.48\textwidth} % '[b]'表示子图的垂直对齐方式(底部对齐), '0.48\textwidth' 表示子图宽度为文本宽度的48%
        \centering % 使子图内的图片居中
        \includegraphics[width=\linewidth]{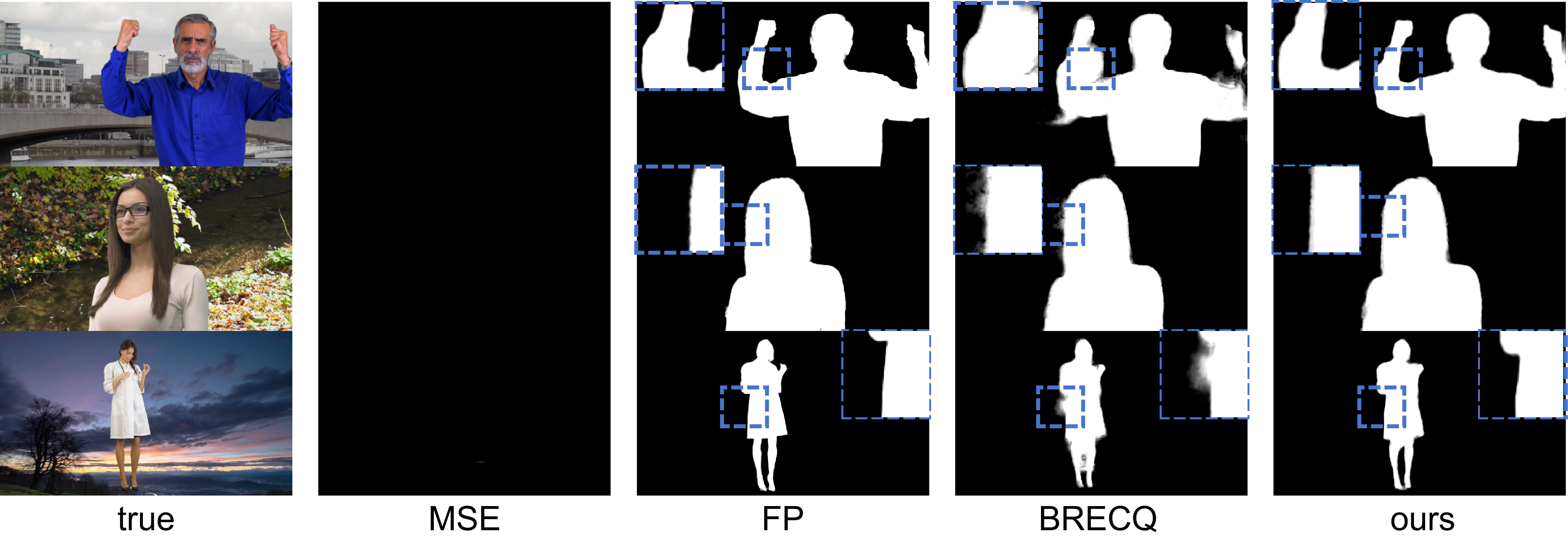} % 图片宽度撑满子图定义的宽度
        \caption{W4A8 Quantization} % 子图a的标题 - 权重4bit，激活8bit
        \label{fig:subfig_w4a8} % 子图a的标签
    \end{subfigure}%
    \hfill% 在两个子图之间添加一些水平弹性空间，并确保此处的换行不产生空格
    % 第二个子图 (b)
    \begin{subfigure}[b]{0.48\textwidth}
        \centering
        \includegraphics[width=\linewidth]{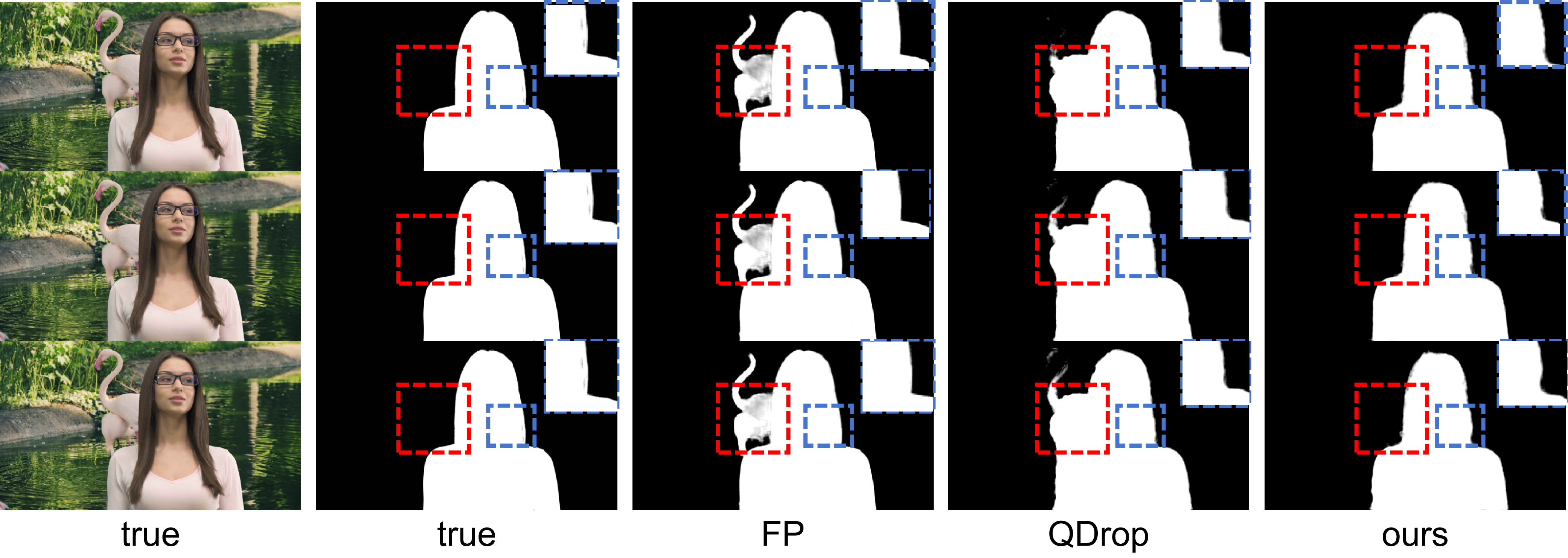}
        \caption{W4A4 Quantization} % 子图b的标题 - 权重4bit，激活4bit
        \label{fig:subfig_w4a4} % 子图b的标签
    \end{subfigure}

    % 整体 Figure 的总标题
    % 中文注释：比较PTQ4VM与基准模型在(a)W4A8和(b)W4A4量化下的效果。我们的方法展示了优越的准确性和视频理解能力。
    \caption{Comparison of PTQ4VM with Ground Truth (true), Full-Precision (FP) RVM, MSE, BRECQ, and QDrop under (a) W4A8 and (b) W4A4 quantization. Our method demonstrates superior accuracy and video understanding capabilities.}
    \label{fig:comparison_ptq4vm_results} % 整个figure的标签
\end{figure}

We also provide visual comparisons. As shown in Figure~\ref{fig:subfig_w4a8}, our training framework enhances matting accuracy, exhibiting better performance on intricate curve and motion details.Figure~\ref{fig:subfig_w4a4} demonstrates the improvement in video semantic understanding due to our framework. Even full-precision models sometimes fail to distinguish similar static background interference, but our model accurately identifies the moving foreground, which also corroborates the guiding role of the OFA component.

% 表格修改：使用 \begin{table} 和 \centering，调整 \multicolumn 和 \cmidrule 模仿示例。
% 它将展示不同方法（例如，FP32 RVM，我们的PTQ RVM）在不同数据集上的多种评估指标结果。
% 表格中的具体数值将由实际实验结果填充。
\begin{table*}[tbp] % 如果是双栏文档中的单栏表格，改为 
  \centering
  \caption{Quantitative comparison of our PTQ method against the FP32 baseline. Results for FLOPs, Params, and metrics are to be filled by the user. Lower values are better for all metrics. MSE for Alpha and FG is reported $\times 10^{-3}$.}
  \label{main} % 建议每次修改后更新label以便区分
  % 使用 \resizebox 解决表格过宽问题。如果是单栏表格，将 \textwidth 改为 \linewidth
  \resizebox{\textwidth}{!}
  {%
    % 为了调试，暂时将数字列改为了普通的 'r' (右对齐)
    % 原计划使用 siunitx 的 S 列，例如: ll S[table-format=2.0] S[table-format=3.1] ...
    % 当前列定义: l l r r r r r r r r r (共11列数据 + 2列描述 = 13列，但Dataset和Method是描述列，数据内容为11列)
    % 应该是 Dataset(l), Method(l), #Bit(r), #FLOPs(r), #Param(r), MAD(r), MSE(r), Grad(r), Conn(r), DTSSD(r), FG MSE(r) -> ll + 9*r
    \begin{tabular}{ll rrrrrrrrr}
      \toprule
      % --- 表头第一行 ---
      % 中文注释：请确保您的代码中，花括号内的文本就是 "Dataset" 和 "Method"，没有 "2*"
      \multirow{2}{*}{Dataset} & \multirow{2}{*}{Method} & {\#Bit} & {\#FLOPs} & {\#Param} & \multicolumn{5}{c}{Alpha ($\alpha$)} & {FG} \\
      % 中文注释：定义Alpha和FG下的细分列规则
      \cmidrule(lr){6-10} \cmidrule(lr){11-11}
      % --- 表头第二行 (单位和指标名称) ---
       &  &  & {(G)$\downarrow$} & {(MB)$\downarrow$} & {MAD$\downarrow$} & {MSE$\downarrow$} & {Grad$\downarrow$} & {Conn$\downarrow$} & {DTSSD$\downarrow$} & {MSE$\downarrow$} \\
      \midrule
      
      % --- VM Dataset ---
      % 中文注释：请确保您的代码中，\shortstack 内的文本就是 "VM \\ 512x288"，没有 "2*"
      \multirow{13}{*}{\begin{tabular}{@{}c@{}}VM \\ 512x288\end{tabular}} % 您需要根据实际情况修改分辨率
      & RVM (FP32 Baseline) & W32A32 & {4.57} & {14.5} & {6.08} & {1.47} & {0.88} & {0.41} & {1.36} & {-} \\ 
      \cmidrule(lr){2-11}
      & MSE               & W8A8  & {1.14} & {3.63} & {6.36} & {1.43} & {1.13} & {0.45} & {1.63} & {-} \\
      & BRECQ               & W8A8  & {1.14} & {3.63} & {6.17} & \textbf{1.27} & {1.05} & {0.42} & {1.76} & {-} \\
      & QDrop               & W8A8  & {1.14} & {3.63} & {6.24} & {1.54} & {0.96} & {0.44} & {1.49} & {-} \\
      & Our PTQ RVM         & W8A8  & {1.14} & {3.63} & \textbf{6.03} & {1.29} & \textbf{0.95} & \textbf{0.41} & \textbf{1.46} & {-} \\
      \cmidrule(lr){2-11}
      & MSE               & W4A8  & {0.76} & {2.42} & {168.22} & {158.09} & {14.25} & {24.34} & {4.53} & {-} \\
      & BRECQ               & W4A8  & {0.76} & {2.42} & {28.67} & {19.94} & {7.47} & {3.84} & {3.35} & {-} \\
      & QDrop               & W4A8  & {0.76} & {2.42} & {11.72} & {5.28} & {3.75} & {1.30} & {2.55} & {-} \\
      & Our PTQ RVM         & W4A8  & {0.76} & {2.42} & \textbf{10.77} & \textbf{4.54} & \textbf{3.49} & \textbf{1.15} & \textbf{2.51} & {-} \\
      \cmidrule(lr){2-11}
      & MSE               & W4A4  & {0.57} & {1.81} & {189.21} & {184.38} & {15.08} & {27.40} & \textbf{3.81} & {-} \\
      & BRECQ               & W4A4  & {0.57} & {1.81} & {168.34} & {161.61} & {15.27} & {24.36} & {5.10} & {-} \\
      & QDrop               & W4A4  & {0.57} & {1.81} & {24.36} & {18.02} & {8.92} & {3.16} & {4.70} & {-} \\
      & Our PTQ RVM         & W4A4  & {0.57} & {1.81} & \textbf{20.33} & \textbf{13.80} & \textbf{7.48} & \textbf{2.57} & {4.63} & {-} \\
      \midrule
      
      % --- D646 Dataset ---
      % 中文注释：请确保您的代码中，\shortstack 内的文本就是 "D646 \\ 512x512"，没有 "2*"
      \multirow{13}{*}{\begin{tabular}{@{}c@{}}D646 \\ 512x512\end{tabular}} % 您需要根据实际情况修改分辨率
      & RVM (FP32 Baseline) & W32A32 & {8.12} & {14.5} & {6.63} & {1.91} & {2.43} & {1.60} & {0.80} & {2.54} \\ % DTSSD 不适用于图像 
      \cmidrule(lr){2-11}
      & MSE & W8A8 & {2.03} & {3.63} & {8.03} & {2.56} & {3.22} & {1.97} & {1.10} & {2.77} \\
      & BRECQ & W8A8 & {2.03} & {3.63} & {7.25} & {2.33} & {2.89} & {1.77} & {1.07} & \textbf{2.53} \\
      & QDrop & W8A8 & {2.03} & {3.63} & {7.19} & \textbf{2.20} & \textbf{2.85} & {1.77} & {0.98} & {2.58} \\
      & Our PTQ RVM & W8A8 & {2.03} & {3.63} & \textbf{7.14} & {2.23} & {2.92} & \textbf{1.76} & \textbf{0.92} & {2.58} \\
      \cmidrule(lr){2-11}
      & MSE & W4A8 & {1.35} & {2.42} & {234.09} & {228.48} & {29.43} & {61.19} & \textbf{1.38} & {26.61} \\
      & BRECQ & W4A8 & {1.35} & {2.42} & {60.67} & {50.88} & {18.22} & {15.98} & {1.94} & {16.56} \\
      & QDrop & W4A8 & {1.35} & {2.42} & {19.93} & {11.89} & {10.35} & {5.28} & {1.62} & \textbf{4.69} \\
      & Our PTQ RVM & W4A8 & {1.35} & {2.42} & \textbf{18.77} & \textbf{11.14} & \textbf{9.94} & \textbf{4.97} & {1.61} & {4.97} \\
      \cmidrule(lr){2-11}
      & MSE & W4A4 & {1.02} & {1.81} & {234.11} & {228.50} & {29.48} & {61.19} & \textbf{1.49} & {11.98} \\
      & BRECQ & W4A4 & {1.02} & {1.81} & {216.46} & {208.53} & {30.24} & {56.64} & {3.77} & {90.92} \\
      & QDrop & W4A4 & {1.02} & {1.81} & {47.91} & {40.15} & {20.85} & {12.60} & {2.36} & {9.13} \\
      & Our PTQ RVM   & W4A4  & {1.02} & {1.81} & \textbf{46.73} & \textbf{38.63} & \textbf{18.54} & \textbf{12.26} & {3.52} & \textbf{8.68} \\
      % \midrule
      
      % --- AIM-500 Dataset ---
      % 中文注释：请确保您的代码中，\shortstack 内的文本就是 "AIM-500 \\ 512x512"，没有 "2*"
      % \multirow{2}{*}{\shortstack{AIM-500 \\ 512x512}} % 您需要根据实际情况修改分辨率
      % & RVM (FP32 Baseline) & W32A32 & {} & {} & {5.09} & {1.49} & {1.88} & {1.24} & {0.69} & {5.02} \\ % DTSSD 不适用于图像
      % & MaxMin & W8A8 & {} & {} & {6.12} & {2.09} & {2.52} & {1.51} & {0.99} & {4.91} \\ % DTSSD 不适用于图像
      % & BRECQ & W8A8 & {} & {} & {5.43} & {1.79} & {2.16} & {1.31} & {0.95} & {5.28}
      % & QDrop & W8A8 & {} & {} & {5.55} & {1.70} & {2.11} & {1.38} & {0.87} & {5.18}
      % & Our PTQ RVM (INT8)  & W8A8  & {} & {} & {} & {} & {} & {} & {-} & {}
      % & MaxMin & W4A8 & {} & {} & {377.06} & {373.03} & {26.59} & {98.69} & {1.04} & {28.85} \\
      % & BRECQ & W4A8 & {} & {} & {108.85} & {97.38} & {19.23} & {28.81} & {2.43} & {26.84}
      % & QDrop & W4A8 & {} & {} & {13.22} & {7.19} & {7.83} & {3.84} & {1.39} & {6.10}
      % & Our PTQ RVM (INT8)  & 8  & {} & {} & {} & {} & {} & {} & {-} & {}
      % & MaxMin & W4A4 & {} & {} & {377.08} & {373.04} & {26.63} & {98.70} & {1.20} & {28.56} \\
      % & BRECQ & W4A4 & {} & {} & {376.06} & {371.74} & {26.98} & {98.44} & {1.60} & {81.22}
      % & QDrop & W8A8 & {} & {} & {33.81} & {27.74} & {16.71} & {8.88} & {3.50} & {11.56}
      % & Our PTQ RVM (INT8)  & 8  & {} & {} & {} & {} & {} & {} & {-} & {} \\
      \bottomrule
    \end{tabular}%
  } % 结束 \resizebox
\end{table*}

\begin{table*}[tbp] % 如果是双栏文档中的单栏表格，改为 \begin{table}[htbp]
  \centering
  \caption{Quantitative comparison of our PTQ method against the FP32 baseline. Results for FLOPs, Params, and metrics are to be filled by the user. Lower values are better for all metrics. MSE for Alpha and FG is reported $\times 10^{-3}$.}
  \label{main2} % 建议每次修改后更新label以便区分
  % 使用 \resizebox 解决表格过宽问题。如果是单栏表格，将 \textwidth 改为 \linewidth
  \resizebox{\textwidth}{!}{%
    % 为了调试，暂时将数字列改为了普通的 'r' (右对齐)
    % 原计划使用 siunitx 的 S 列，例如: ll S[table-format=2.0] S[table-format=3.1] ...
    % 当前列定义: l l r r r r r r r r r (共11列数据 + 2列描述 = 13列，但Dataset和Method是描述列，数据内容为11列)
    % 应该是 Dataset(l), Method(l), #Bit(r), #FLOPs(r), #Param(r), MAD(r), MSE(r), Grad(r), Conn(r), DTSSD(r), FG MSE(r) -> ll + 9*r
    \begin{tabular}{ll rrrrrrrrr}
      \toprule
      % --- 表头第一行 ---
      % 中文注释：请确保您的代码中，花括号内的文本就是 "Dataset" 和 "Method"，没有 "2*"
      \multirow{2}{*}{Dataset} & \multirow{2}{*}{Method} & {\#Bit} & {\#FLOPs} & {\#Param} & \multicolumn{5}{c}{Alpha ($\alpha$)} & {FG} \\
      % 中文注释：定义Alpha和FG下的细分列规则
      \cmidrule(lr){6-10} \cmidrule(lr){11-11}
      % --- 表头第二行 (单位和指标名称) ---
       &  &  & {(G)$\downarrow$} & {(MB)$\downarrow$} & {MAD$\downarrow$} & {MSE$\downarrow$} & {Grad$\downarrow$} & {Conn$\downarrow$} & {DTSSD$\downarrow$} & {MSE$\downarrow$} \\
      \midrule
      
      % --- VM Dataset ---
      % 中文注释：请确保您的代码中，\shortstack 内的文本就是 "VM \\ 512x288"，没有 "2*"
      \multirow{8}{*}{\begin{tabular}{@{}c@{}}VM \\ 512x288\end{tabular}} % 您需要根据实际情况修改分辨率
      % & BRECQ               & W8A8  & {1.14} & {3.63} & {6.17} & \textbf{1.27} & {1.05} & {0.42} & {1.76} & {-} \\
      % & QDrop               & W8A8  & {1.14} & {3.63} & {6.24} & {1.54} & {0.96} & {0.44} & {1.49} & {-} \\
      % & QDrop+LCW             & W8A8  & {1.14} & {3.63} & {6.18} & {1.46} & {0.96} & {0.43} & {1.48} & {-} \\
      % \cmidrule(lr){2-11}
      & BRECQ               & W4A8  & {0.76} & {2.42} & {28.67} & {19.94} & {7.47} & {3.84} & {3.35} & {-} \\
      & BRECQ+GAC             & W4A8  & {0.76} & {2.42} & {14.91} & {7.21} & {3.37} & {1.73} & {2.50} & {-} \\
      & BRECQ+GAC+OFA             & W4A8  & {0.76} & {2.42} & {13.18} & {6.78} & {3.25} & {1.48} & {2.59} & {-} \\
      \cmidrule(lr){2-11}
      & QDrop               & W4A8  & {0.76} & {2.42} & {11.72} & {5.28} & {3.75} & {1.30} & {2.55} & {-} \\
      & QDrop+GAC               & W4A8  & {0.76} & {2.42} & {10.98} & {4.43} & {3.36} & {1.17} & {2.46} & {-} \\
      & QDrop+GAC+OFA             & W4A8  & {0.76} & {2.42} & {10.77} & {4.54} & {3.49} & {1.15} & {2.51} & {-} \\
      \cmidrule(lr){2-11}
      & BRECQ               & W4A4  & {0.57} & {1.81} & {168.34} & {161.61} & {15.27} & {24.36} & {5.10} & {-} \\
      & BRECQ+GAC              & W4A4  & {0.57} & {1.81} & {50.75} & {39.84} & {10.44} & {7.11} & {8.01} & {-} \\
      & BRECQ+GAC+OFA              & W4A4  & {0.57} & {1.81} & {46.16} & {27.29} & {7.29} & {5.17} & {3.15} & {-} \\
      \cmidrule(lr){2-11}
      & QDrop               & W4A4  & {0.57} & {1.81} & {24.36} & {18.02} & {8.92} & {3.16} & {4.70} & {-} \\
      & QDrop+GAC        & W4A4  & {0.57} & {1.81} & {22.01} & {11.85} & {6.90} & {2.80} & {3.96} & {-} \\
      & QDrop+GAC+OFA        & W4A4  & {0.57} & {1.81} & {20.33} & {13.80} & {7.48} & {2.57} & {4.63} & {-} \\
      % \midrule      
     
      \bottomrule
    \end{tabular}%
  } % 结束 \resizebox
\end{table*}

\subsection{Ablation Studies}
% **全局仿射校准 (GAC) 的有效性与通用性分析**
\paragraph{Effectiveness and Generality Analysis of Global Affine Calibration (GAC)}
% 为验证层级全局仿射校准 (GAC) 策略的有效性及其作为通用后处理模块的潜力，本研究将GAC模块独立应用于两种先进的PTQ算法BRECQ与QDrop之上。
We apply the GAC module independently to two state-of-the-art PTQ algorithms, BRECQ and QDrop.
% 具体而言，我们首先获得由BRECQ和QDrop在不同位宽下生成的量化模型，随后施加GAC进行权重微调，其优化目标与我们完整框架中第二阶段的设定保持一致。
Specifically, we first obtain the quantized models generated by BRECQ and QDrop at different bit-widths, and then apply GAC for further fine-tuning, with the optimization objective consistent with that of Stage 2 in our complete framework.

% 如表2 (\ref{tab:gac_effectiveness_comparison}) 所示，GAC在低位宽设定（尤其是W4A4）下，显著提升了BRECQ和QDrop在各项指标上的性能。
As shown in Table~\ref{main2}, GAC significantly enhances the performance of both BRECQ and QDrop across various metrics under low bit-width settings, particularly for W4A4.

% 值得注意的是，GAC对BRECQ的性能增益尤为显著。应用GAC后，BRECQ的各项指标均得到大幅改善，使其性能达到了与QDrop（在某些指标上甚至接近或优于未经GAC处理的QDrop）相当的水平。
Notably, the performance gain from GAC is particularly significant for BRECQ. After applying GAC, nearly all metrics for BRECQ improve substantially, bringing its performance to a level comparable with QDrop without GAC.
% 此现象颇具启发性，为理解QDrop等通过模拟量化噪声进行优化的方法提供了新视角。
This phenomenon is quite insightful, offering a new perspective for understanding methods like QDrop that optimize by simulating quantization noise.
% QDrop在训练中通过随机扰动学习对统计偏差具有鲁棒性的权重及量化参数，而GAC则通过简洁的线性变换直接对量化引入的统计偏差进行全局补偿。
QDrop learns weights and quantization parameters robust to statistical deviations through random perturbations during training, whereas GAC directly compensates for the statistical deviations introduced by quantization via a simple linear transformation globally.

% GAC的成功应用证明，一个简洁高效的后处理校准模块能够有效提升现有复杂PTQ方法的性能上限，尤其适用于校准数据有限且难以进行重训练或精细统计调整的场景。

% \begin{figure}[htbp] % Floating parameter suggestions
%     \includegraphics[width=0.4\textwidth]{04.png} % Adjust width to fit your document
%     \caption{This is a caption for the line graph, describing its content.} % English caption
%     \label{fig:my_line_graph} % Label for cross-referencing
% \end{figure}
\paragraph{Effectiveness of the Optical Flow-Assisted (OFA) Component}

We investigate the potential benefits of the OFA component for the second-stage calibration of existing PTQ methods. As indicated in Table~\ref{main2}, when the OFA component is integrated into the second-stage calibration process for both BRECQ and QDrop, further improvements in accuracy are observed for both methods. This suggests that the temporal priors provided by OFA can effectively guide the optimization within our own framework.

\paragraph{Limitation}Finally, our method cannot fully achieve the detail-capturing capability of full-precision models, and the degradation of model representational capacity remains a challenge under extremely low bit-widths (e.g., 1-2 bits).

% 除了定量评估外，我们还提供了可视化结果，以直观地比较我们的量化模型与全精度基线模型在不同场景下的抠图质量和时间一致性。
\section{Conclusion}

% 本文据我们所知，首次针对视频抠图任务提出了一种高效的训练后量化（PTQ）框架。
To the best of our knowledge, this paper presents the first effective Post-Training Quantization (PTQ) framework specifically tailored for the video matting task.
% 我们提出了一种通用的多阶段量化策略，该策略首先通过逐块优化进行初始量化，随后采用层级线性权重校准来精细调整量化参数。
We have proposed a general multi-stage quantization strategy that first performs initial quantization via block-wise optimization, followed by a layer-wise linear weight calibration to refine quantization parameters.
% 此外，我们创新性地引入了光流辅助（OFA）组件，该组件不仅显著提升了量化模型在长视频序列上的时序一致性，还增强了其视频语义理解能力。
Furthermore, we innovatively introduced an Optical Flow-Assisted (OFA) component, which not only significantly enhances the temporal consistency of the quantized model over long video sequences but also improves its video semantic understanding capabilities.
% 实验结果表明，我们的方法能够在大幅降低模型计算量和存储需求的同时，保持与全精度模型相媲美的抠图质量，甚至在极低比特下仍表现出优越的鲁棒性和泛化能力。
Experiments demonstrate that our method can maintain matting quality comparable to full-precision models while substantially reducing model computation and storage requirements, exhibiting superior robustness and generalization even at very low bit-widths.
% 这项工作为视频抠图模型在资源受限设备上的实际部署提供了可行的解决方案，并为未来复杂视频处理任务的PTQ研究提供了有价值的见解。
This work offers a viable solution for the practical deployment of video matting models on resource-constrained devices and provides insights for future PTQ research in complex video processing tasks.
This methodology also exemplifies an effective application of optical flow, showcasing its utility in regularizing the fine-tuning of quantized models for temporally coherent video processing.

\clearpage

\label{others}

% These instructions apply to everyone.

% \section*{References}

% References follow the acknowledgments in the camera-ready paper. Use unnumbered first-level heading for
% the references. Any choice of citation style is acceptable as long as you are
% consistent. It is permissible to reduce the font size to \verb+small+ (9 point)
% when listing the references.
% Note that the Reference section does not count towards the page limit.
% \medskip

% {
% \small

% [1] Alexander, J.A.\ \& Mozer, M.C.\ (1995) Template-based algorithms for
% connectionist rule extraction. In G.\ Tesauro, D.S.\ Touretzky and T.K.\ Leen
% (eds.), {\it Advances in Neural Information Processing Systems 7},
% pp.\ 609--616. Cambridge, MA: MIT Press.

% [2] Bower, J.M.\ \& Beeman, D.\ (1995) {\it The Book of GENESIS: Exploring
%   Realistic Neural Models with the GEneral NEural SImulation System.}  New York:
% TELOS/Springer--Verlag.

% [3] Hasselmo, M.E., Schnell, E.\ \& Barkai, E.\ (1995) Dynamics of learning and
% recall at excitatory recurrent synapses and cholinergic modulation in rat
% hippocampal region CA3. {\it Journal of Neuroscience} {\bf 15}(7):5249-5262.
% }

\bibliography{references}
\bibliographystyle{plain}
%%%%%%%%%%%%%%%%%%%%%%%%%%%%%%%%%%%%%%%%%%%%%%%%%%%%%%%%%%%%

\appendix

\section{Technical Appendices and Supplementary Material}
\subsection{Analysis of Block-wise Initial Quantization (BIQ) Convergence}
% 中文标题：BIQ优化粒度与收敛性分析

% B.1 引言与动机
% 正如主论文第X.X节（BIQ方法介绍部分）所讨论的，优化粒度的选择对后训练量化的最终性能至关重要。
As discussed in Section~{3.2} of the main paper, the choice of optimization granularity is critical to the final performance of Post-Training Quantization (PTQ).
% 本节通过展示不同量化位宽下，逐块优化与朴素全网络量化在Alpha绝对差异均值 (MAD) 上的收敛曲线，为这一选择提供实验支撑。
This section provides experimental support for this choice by presenting the convergence curves of Alpha error) for block-wise optimization versus naive full-network quantization under different bit-width settings.

\begin{figure}[htbp]
    \centering
    % W4A4 子图
    \begin{subfigure}[b]{0.48\textwidth} % 调整宽度以适应并排两个图
        \centering
        \includegraphics[width=\linewidth]{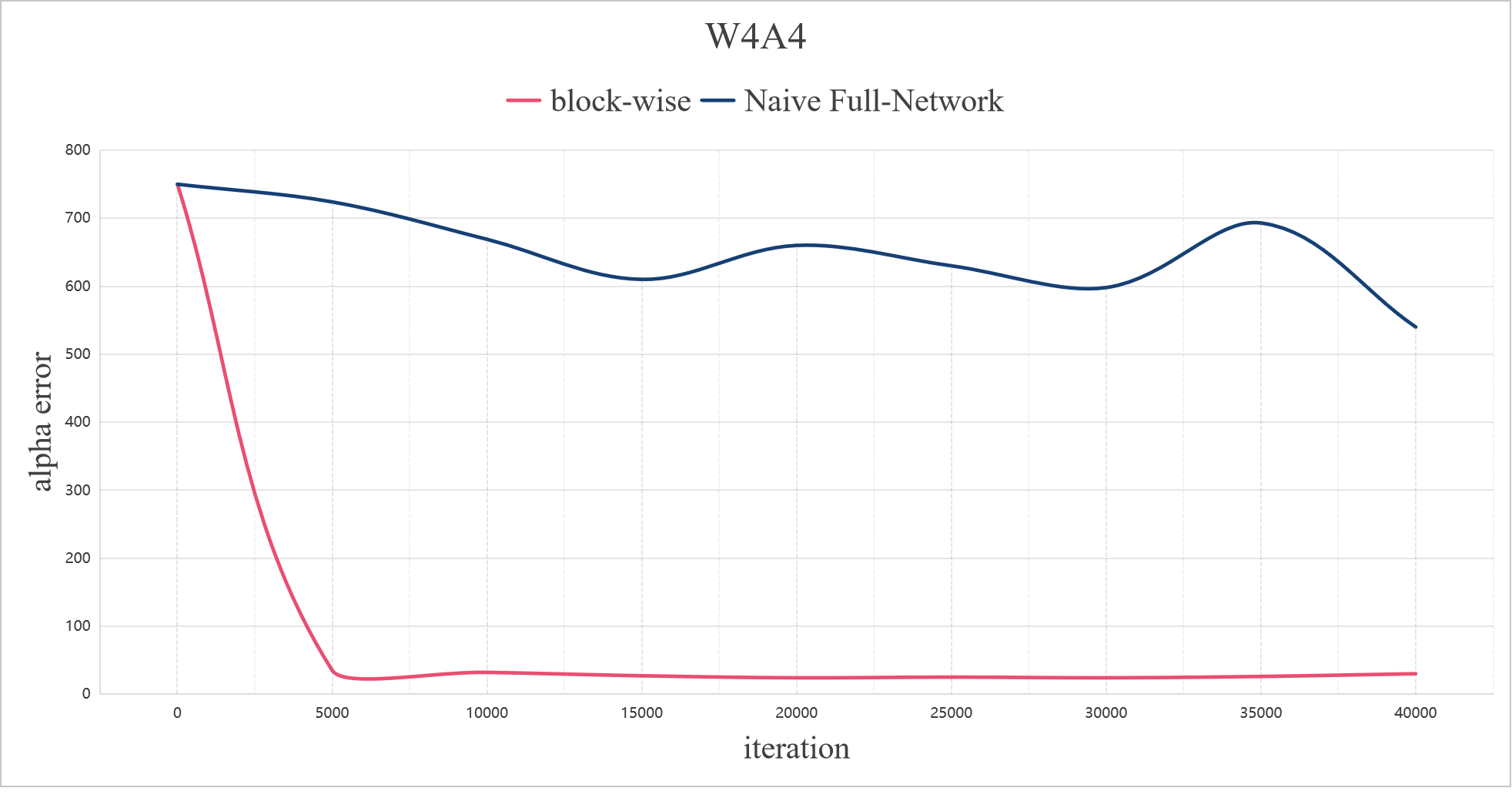} % W4A4收敛图
        \caption{W4A4 Setting}
        \label{fig:sup_conv_w4a4_sub}
    \end{subfigure}%
    \hfill % 水平间距
    % W4A8 子图
    \begin{subfigure}[b]{0.48\textwidth} % 调整宽度以适应并排两个图
        \centering
        \includegraphics[width=\linewidth]{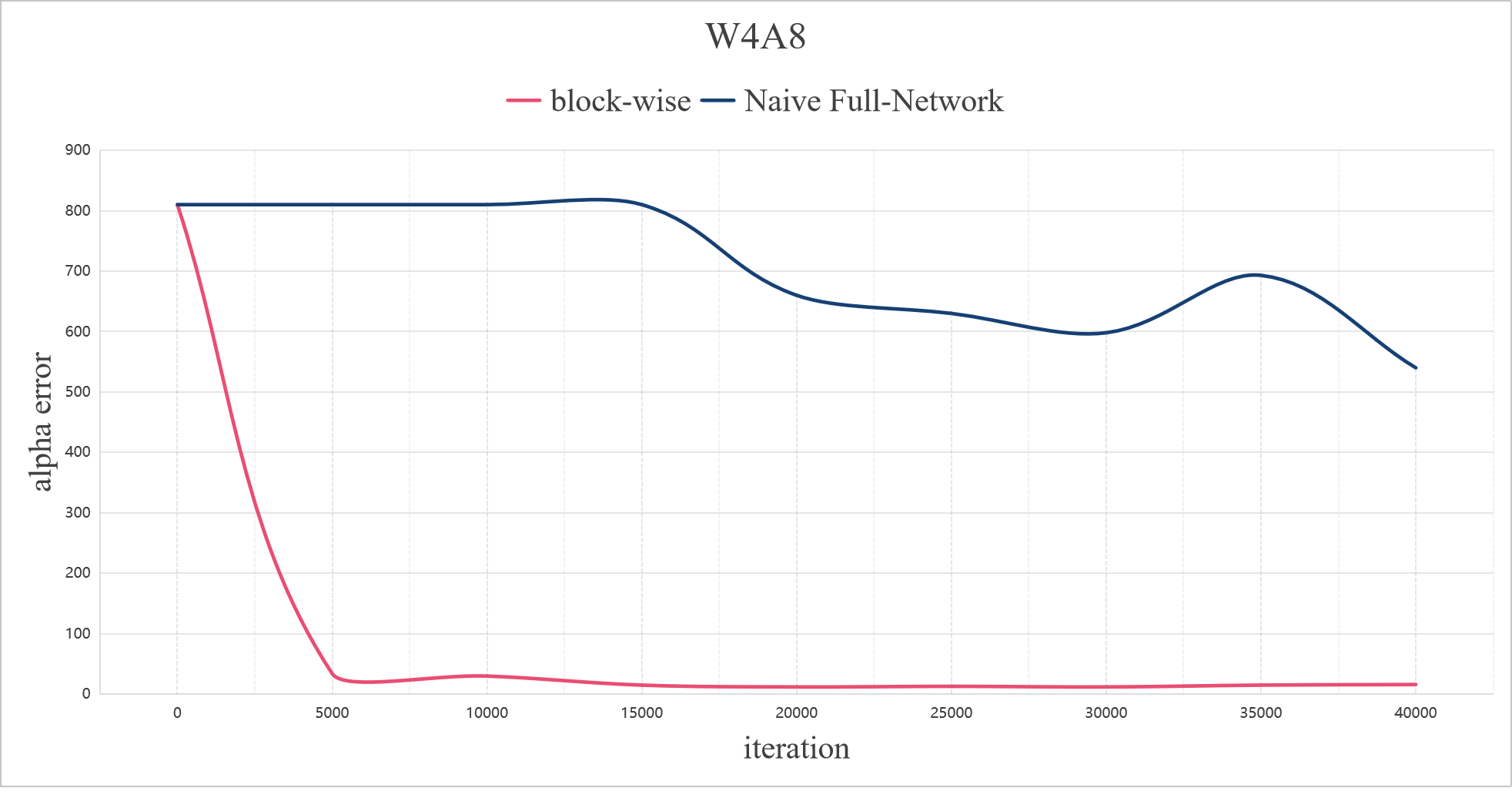} % W4A8收敛图
        \caption{W4A8 Setting}
        \label{fig:sup_conv_w4a8_sub}
    \end{subfigure}
    % 中文图注：图S3: 不同量化设置下，逐块初始量化 (BIQ) 与朴素全网络量化的Alpha MAD收敛曲线对比。(a) W4A4, (b) W4A8。(请在图中将图例的 "model-wise" 修改为 "Naive Full-Network" 或类似表述)
    \caption{Convergence comparison of Alpha error for Block-wise Initial Quantization (BIQ) versus Naive Full-Network Quantization under different settings: (a) W4A4 and (b) W4A8. Evaluations are performed every 5000 iterations, and the curves are smoothed for clarity.}
    \label{fig:sup_conv_combined}
\end{figure}

% B.2 不同位宽下的收敛性比较
\paragraph{Convergence Comparison under Various Bit-widths}
% 我们比较了在W4A4和W4A8两种不同权重-激活位宽设置下，我们提出的逐块初始量化 (BIQ) 方法与一种朴素的全网络直接量化方法（作为对比基线，其尝试一次性优化整个网络的量化参数以最小化与全精度输出的MAD）的收敛过程。
We compare the convergence process of our proposed Block-wise Initial Quantization (BIQ) method against a naive full-network direct quantization approach (which attempts to optimize quantization parameters for the entire network at once to minimize MSE against the full-precision output, serving as a baseline for comparison) under two different weight-activation bit-width settings: W4A4 and W4A8.
% 优化目标均为最小化块输出（对于BIQ）或网络最终Alpha输出（对于全网络量化）与全精度对应输出之间的绝对差异均值。
The optimization objective for both is to minimize the Mean Square Error (MSE) between the block output (for BIQ) or the final network alpha output (for full-network quantization) and their full-precision counterparts.
% 图S3展示了这两种设置下的Alpha MAD损失随迭代次数的变化。
Figure~\ref{fig:sup_conv_combined} illustrates the Alpha error, evaluated on the test set, versus the number of iterations for these two settings.

% 从这些收敛曲线中，我们可以清晰地观察到：
From these convergence curves (Figure~\ref{fig:sup_conv_combined}), we can clearly observe:
\begin{itemize}
    % * **BIQ的有效收敛与朴素全网络量化的收敛困难**：在所有测试的位宽下（W4A4, W4A8），我们的逐块初始量化 (BIQ) 方法均展现出快速且有效的收敛。BIQ的误差曲线在较少的迭代次数内就迅速下降并稳定在较低的水平。相比之下，朴素全网络量化方法的误差曲线几乎没有表现出明显的收敛趋势，其误差值持续处于较高水平，表明直接对整个复杂网络进行优化难以找到有效的量化解。
    \item \textbf{Effective Convergence of BIQ versus Difficulty of Naive Full-Network Quantization} Across the tested bit-widths (W4A4 and W4A8), our Block-wise Initial Quantization (BIQ) method exhibits rapid and effective convergence. The error curve for BIQ drops quickly and stabilizes at a low level within a smaller number of iterations. In contrast, the error curve for the naive full-network quantization method shows little to no significant convergence trend, with its error values remaining persistently high, indicating the difficulty of finding an effective quantization solution by directly optimizing the entire complex network.
    % * **BIQ的最终性能优势**：由于其有效的收敛特性，BIQ最终收敛到的Alpha MAD值显著低于朴素全网络量化方法所能达到的水平（如果后者可被视为收敛的话）。这表明通过逐块优化，我们能够为量化参数找到一个远优的初始解，更有效地捕获了局部依赖关系，避免了因一次性优化整个复杂网络而导致的优化停滞或次优解问题。
    \item \textbf{Superior Final Performance of BIQ} Due to its effective convergence, BIQ achieves a final Alpha MAD value significantly lower than what the naive full-network quantization method can reach (if the latter can be considered to have converged at all). This indicates that by optimizing block by block, we can find a far superior initial solution for the quantization parameters, more effectively capturing local dependencies and avoiding the optimization stagnation or sub-optimal solutions often encountered when attempting to optimize the entire complex network at once.
\end{itemize}
\subsection{Analysis of Affine Calibration Parameter Distributions in GAC} % 将原来的section降为subsection
% 为进一步理解我们的全局仿射校准 (GAC) 策略在不同量化位宽（W4A4, W4A8, W8A8）下对模型性能的提升机制，本节详细分析了GAC阶段学习到的层级仿射变换参数——平移因子 $\beta_i$ 和缩放因子 $\gamma_i$——的分布特性。
To further understand the mechanism by which our Global Affine Calibration (GAC) strategy enhances model performance under various quantization bit-widths (W4A4, W4A8, W8A8), this section provides a detailed analysis of the distribution characteristics of the layer-wise affine transformation parameters learned during the GAC stage: the shift factor $\beta_i$ and the scaling factor $\gamma_i$.
% 理想情况下，如果初始量化（例如，经过我们第一阶段BIQ处理后，或应用其他PTQ方法后）已经完美地校准了所有统计偏差，那么学习到的 $\beta_i$ 应接近0，$\gamma_i$ 应接近1。
Ideally, if the initial quantization stage (e.g., after our first-stage BIQ, or after applying other PTQ methods) had perfectly corrected all statistical deviations, the learned $\beta_i$ would be close to 0 and $\gamma_i$ close to 1.
% 本分析旨在揭示GAC实际学习到的参数与这些理想值的偏离程度，从而阐明GAC对初始量化模型的具体补偿作用。
This analysis aims to reveal the extent to which the parameters actually learned by GAC deviate from these ideal values, thereby elucidating the specific compensatory role of GAC for initially quantized models.

\begin{figure}[htbp]
    \centering
    % W4A4 子图
    \begin{subfigure}[b]{0.32\textwidth}
        \centering
        \includegraphics[width=\linewidth]{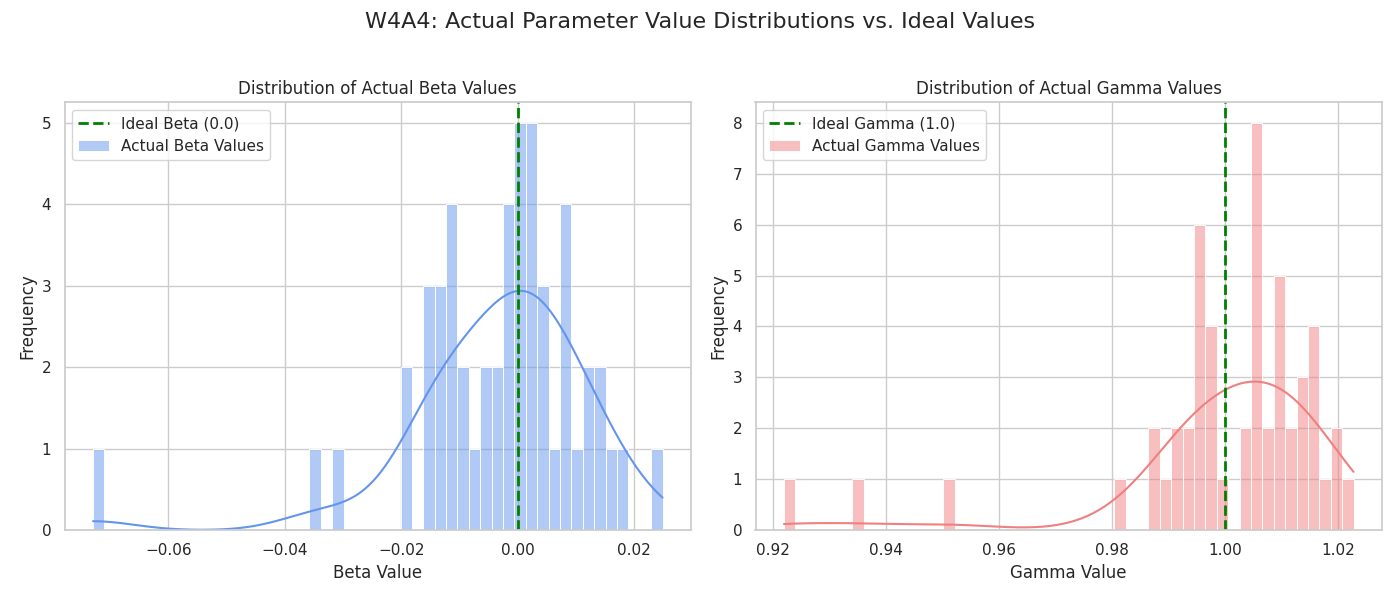} % 替换为您的W4A4图片
        \caption{W4A4 Setting}
        \label{fig:sup_hist_w4a4}
    \end{subfigure}%
    \hfill % 水平间距
    % W4A8 子图
    \begin{subfigure}[b]{0.32\textwidth}
        \centering
        \includegraphics[width=\linewidth]{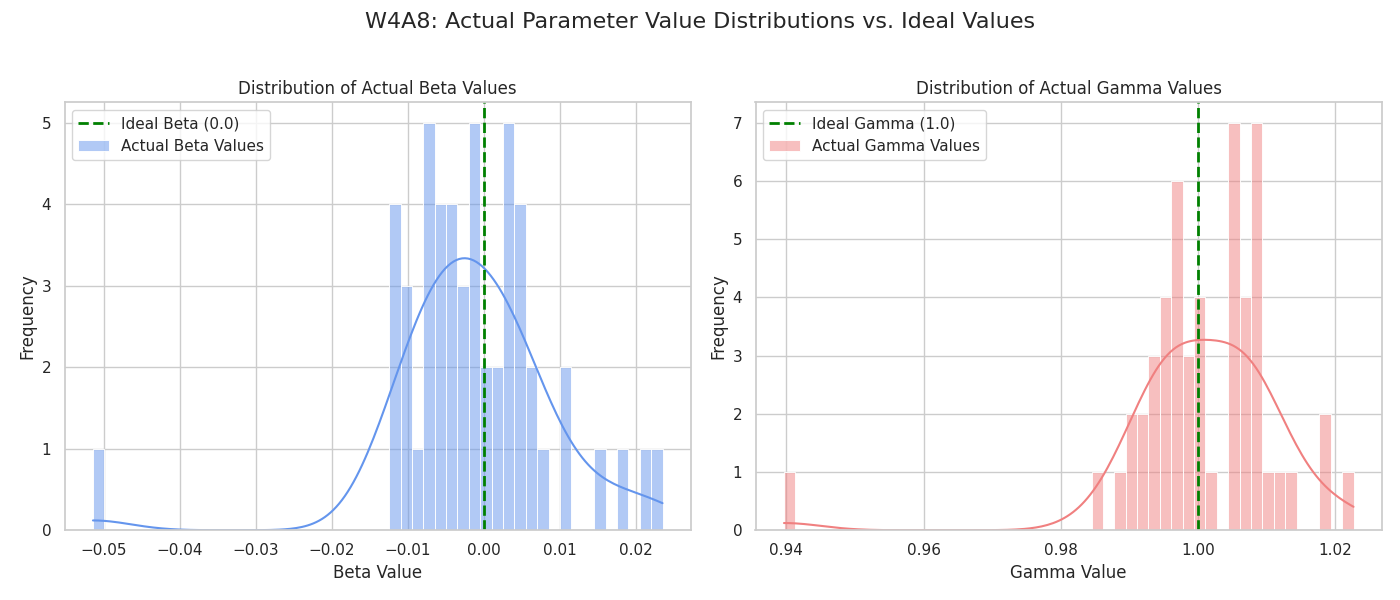} % 替换为您的W4A8图片
        \caption{W4A8 Setting}
        \label{fig:sup_hist_w4a8}
    \end{subfigure}%
    \hfill % 水平间距
    % W8A8 子图
    \begin{subfigure}[b]{0.32\textwidth}
        \centering
        \includegraphics[width=\linewidth]{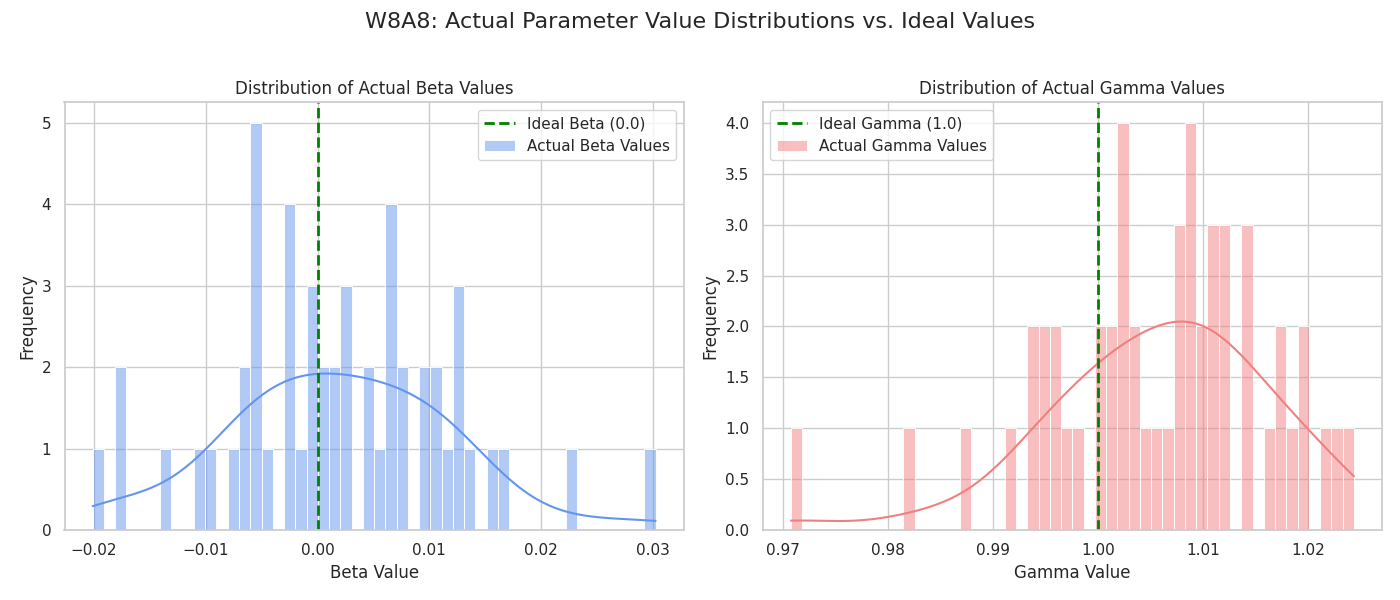} % 替换为您的W8A8图片
        \caption{W8A8 Setting}
        \label{fig:sup_hist_w8a8}
    \end{subfigure}
    % 中文图注：图S1: 不同量化设置下学习到的仿射校准参数 $\beta$ (左列) 和 $\gamma$ (右列，假设每张子图内部包含beta和gamma) 的分布直方图。(a) W4A4, (b) W4A8, (c) W8A8。绿色虚线分别表示理想的 $\beta=0$ 和 $\gamma=1$。
    \caption{Histograms of learned affine calibration parameters $\beta$ and $\gamma$ (each subfigure typically shows distributions for both $\beta$ and $\gamma$) under different quantization settings: (a) W4A4, (b) W4A8, and (c) W8A8. The ideal $\beta=0$ and $\gamma=1$ are typically marked for reference within each panel of the subfigures.}
    \label{fig:sup_param_histograms_combined}
\end{figure}

\begin{figure}[htbp]
    \centering
    % W4A4 子图
    \begin{subfigure}[b]{0.32\textwidth}
        \centering
        \includegraphics[width=\linewidth]{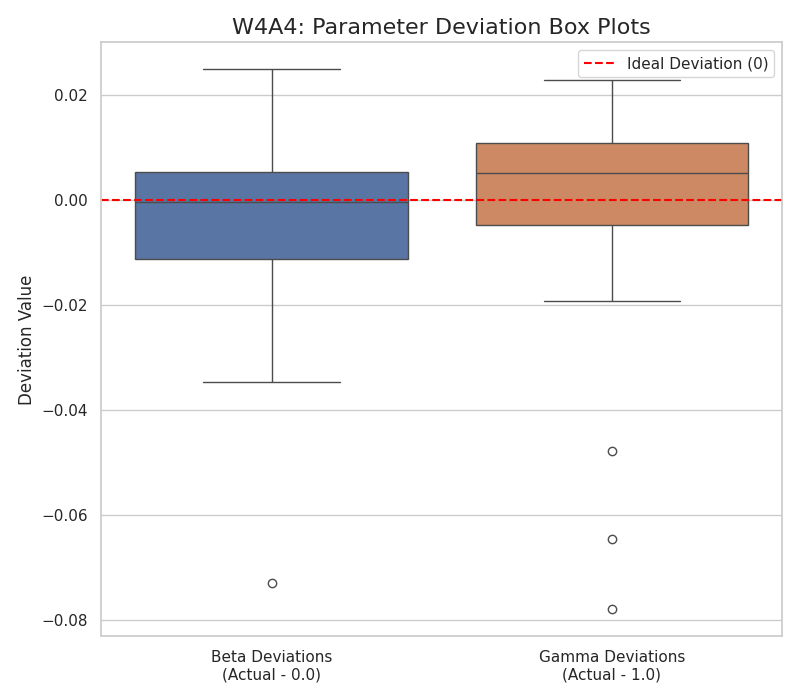} % 替换为您的W4A4箱型图图片
        \caption{W4A4 Deviations}
        \label{fig:sup_boxplot_w4a4}
    \end{subfigure}%
    \hfill % 水平间距
    % W4A8 子图
    \begin{subfigure}[b]{0.32\textwidth}
        \centering
        \includegraphics[width=\linewidth]{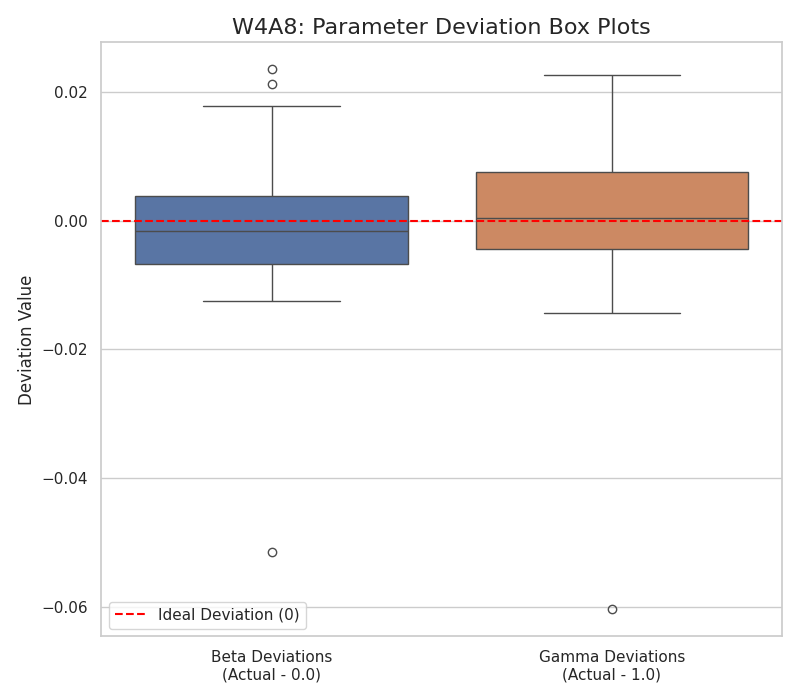} % 替换为您的W4A8箱型图图片
        \caption{W4A8 Deviations}
        \label{fig:sup_boxplot_w4a8}
    \end{subfigure}%
    \hfill % 水平间距
    % W8A8 子图
    \begin{subfigure}[b]{0.32\textwidth}
        \centering
        \includegraphics[width=\linewidth]{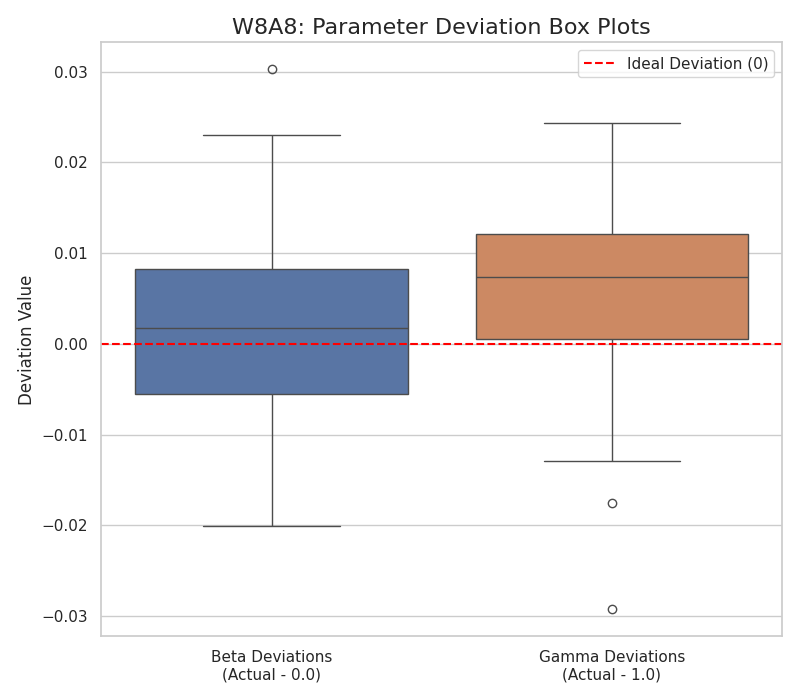} % 替换为您的W8A8箱型图图片
        \caption{W8A8 Deviations}
        \label{fig:sup_boxplot_w8a8}
    \end{subfigure}
    % 中文图注：图S2: 不同量化设置下，学习到的仿射校准参数 $\beta$ 和 $\gamma$ 相对于理想值 (0和1) 的偏差分布箱型图。(a) W4A4, (b) W4A8, (c) W8A8。
    \caption{Box plots of deviations for learned affine calibration parameters $\beta$ (from 0) and $\gamma$ (from 1) under different quantization settings: (a) W4A4, (b) W4A8, and (c) W8A8.}
    \label{fig:sup_param_boxplots_combined}
\end{figure}

% A.2 学习到的仿射参数可视化
\paragraph{Visualization of Learned Affine Parameters}
% 图S1和图S2分别展示了在不同量化设置下，应用于RVM模型各卷积层后学习得到的 $\beta_i$ 和 $\gamma_i$ 参数的实际值分布直方图以及它们与理想值（$\beta=0, \gamma=1$）的偏差箱型图。
Figures~\ref{fig:sup_param_histograms_combined} and \ref{fig:sup_param_boxplots_combined} respectively illustrate the distribution histograms of the actual $\beta_i$ and $\gamma_i$ parameter values learned for each convolutional layer of the RVM model, and the box plots of their deviations from the ideal values ($\beta=0, \gamma=1$), under W4A4, W4A8, and W8A8 quantization settings.

% A.3 参数分布与偏差分析
\paragraph{Analysis of Parameter Distributions and Deviations}
% 图S1和S2共同揭示了学习到的仿射校准参数 $\beta_i$ 和 $\gamma_i$ 的分布特性及其与理想值的偏差。从图中可以客观地观察到，在所有测试的位宽下（W4A4, W4A8, W8A8），学习到的参数均呈现出与理想值（$\beta_i=0, \gamma_i=1$）的偏离。这种偏离在较低位宽（如W4A4）下尤为明显，表现为参数分布更分散，且偏差的绝对值范围更大。
Figures~\ref{fig:sup_param_histograms_combined} and \ref{fig:sup_param_boxplots_combined} collectively reveal the distribution characteristics of the learned affine calibration parameters, $\beta_i$ and $\gamma_i$, and their deviations from ideal values. It is objectively observed from these figures that across all tested bit-widths (W4A4, W4A8, and W8A8), the learned parameters exhibit deviations from their ideal values of $\beta_i=0$ and $\gamma_i=1$. Such deviations are particularly pronounced at lower bit-widths, such as W4A4, where the parameter distributions are more dispersed and the absolute range of deviations is larger.

% 这些观察到的参数偏离现象有力地印证了即使在初始量化阶段（如BIQ）之后，网络权重表示的统计特性（包括均值和尺度）仍存在显著的残余偏差。GAC方法通过学习非零的平移因子 $\beta_i$ 和非单位的缩放因子 $\gamma_i$，正是针对这些统计偏差进行补偿。参数偏离在低位宽下更为显著，进一步凸显了GAC在量化引入较大失真时进行校准的重要性和有效性，从而解释了其对模型性能恢复的关键作用。
These observed parameter deviations strongly corroborate the presence of significant residual statistical alterations (including both mean shifts and scale changes) in the weight representations after the initial quantization stage, even when advanced strategies like BIQ are employed. The GAC method, by learning non-zero shift factors $\beta_i$ and non-unity scaling factors $\gamma_i$, specifically compensates for these statistical discrepancies. The more pronounced deviations at lower bit-widths further underscore the increased importance and efficacy of GAC in calibrating for larger distortions introduced by quantization, thereby explaining its crucial role in model performance recovery.

% A.4 讨论 (保持不变或根据新的分析结构微调)
\paragraph{Discussion}
% 上述分析表明，即使在采用了如BIQ这样的先进初始量化策略后，网络中各层的权重在经过量化后，其统计特性（均值和尺度）与理想状态仍存在差异。
The preceding analysis demonstrates that even after employing advanced initial quantization strategies like BIQ, the statistical properties (mean and scale) of the quantized weights in each layer of the network still differ from an ideal state (where no further affine correction would be needed).
% 全局仿射校准 (GAC) 阶段通过学习层级的平移因子 $\beta_i$ 和缩放因子 $\gamma_i$，有效地对这些残余的统计偏差进行了补偿。
The Global Affine Calibration (GAC) stage effectively compensates for these residual statistical deviations by learning layer-wise shift factors $\beta_i$ and scaling factors $\gamma_i$.
% 这种补偿对于低位宽量化尤为重要，是GAC能够显著提升PTQ模型性能的关键原因之一。
This compensation is particularly crucial for low-bit quantization and is one of the key reasons GAC can significantly enhance the performance of PTQ models.
% 这些学习到的参数分布也反过来印证了在PTQ流程中进行精细统计校准的必要性和有效性。
The distributions of these learned parameters, in turn, corroborate the necessity and effectiveness of performing fine-grained statistical calibration within the PTQ pipeline.

\subsection{Effectiveness of the Optical Flow-Assisted (OFA) Component in Calibration}
% 中文标题：OFA组件在校准过程中的有效性分析

% C.1 引言与实验设置
\paragraph{Experimental Setup}
% 为进一步探究光流辅助 (OFA) 组件在第二阶段校准过程中的具体作用，我们进行了一项对比实验。
To further investigate the specific role of the Optical Flow-Assisted (OFA) component during the second-stage calibration process, we conducted a comparative experiment.
% 该实验在W4A4量化设置下，比较了包含OFA损失项 ($\mathcal{L}_{\text{OFA}}$) 进行联合优化与不包含OFA损失项（即仅使用 $\mathcal{L}_{\alpha}$）进行优化的逐帧平均Alpha误差。
This experiment, under the W4A4 quantization setting, compares the per-frame average Alpha error when performing joint optimization including the OFA loss term ($\mathcal{L}_{\text{OFA}}$) versus optimization using only the $\mathcal{L}_{\alpha}$ loss (i.e., without OFA).
% 实验在VM视频数据集的校准集上进行，逐帧记录了Alpha误差。
The experiment was conducted on the test dataset of the VM video dataset, with Alpha errors recorded frame by frame.

\begin{figure}[htbp]
    \centering
    \includegraphics[width=0.8\textwidth]{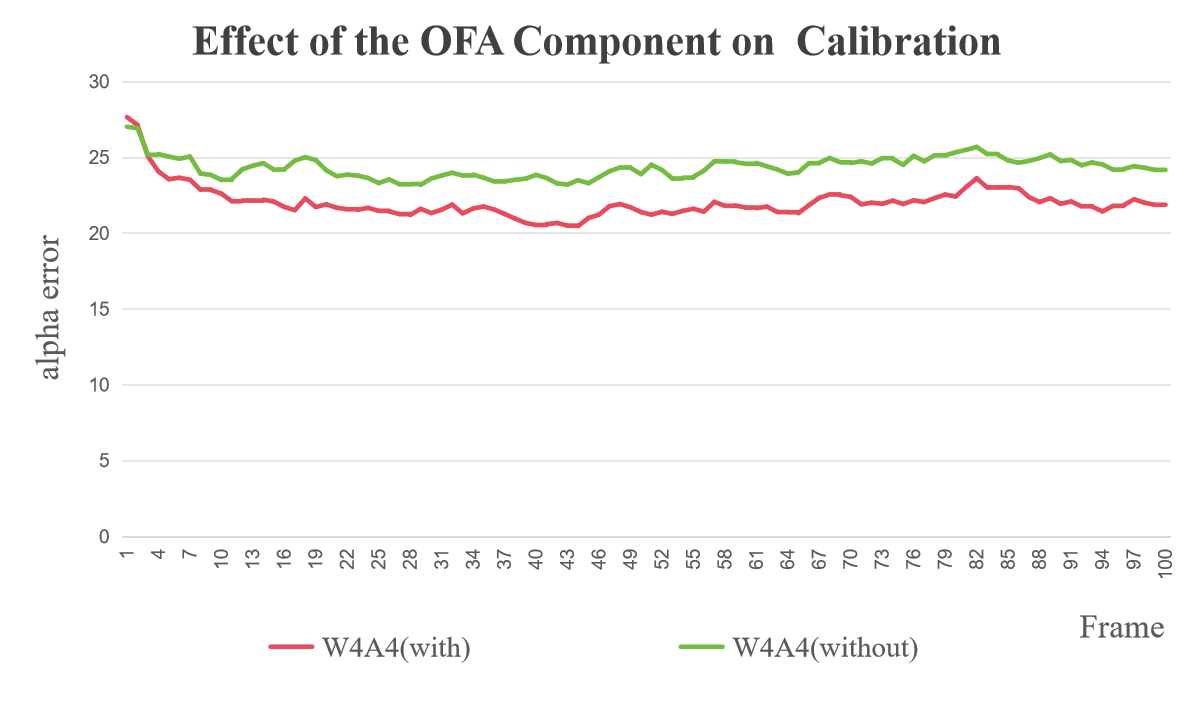} % 替换为您的OFA效果对比图
    % 中文图注：图S4: W4A4量化下，有OFA组件（红色曲线）与无OFA组件（绿色曲线）参与校准的逐帧平均Alpha误差对比。
    \caption{Per-frame average Alpha error comparison for W4A4 quantization with (red curve) and without (green curve) the OFA component on the test dataset.}
    \label{fig:sup_ofa_effect}
\end{figure}

% C.2 逐帧Alpha误差对比分析
\paragraph{Per-Frame Alpha Error Comparison and Analysis}
% 图S4展示了在W4A4量化设置下，有OFA组件和没有OFA组件参与校准时，模型在校准集视频序列上的逐帧平均Alpha误差曲线。
Figure~\ref{fig:sup_ofa_effect} illustrates the per-frame average Alpha error curves on the test dataset video sequences for models calibrated with and without the OFA component under the W4A4 quantization setting, with identical parameters used for the BIQ and GAC stages in these experiments to ensure a fair comparison.

% 从图~\ref{fig:sup_ofa_effect}中可以观察到，在初始几帧，包含OFA组件（红色曲线）与不包含OFA组件（绿色曲线）的模型表现出相似的Alpha误差。然而，随着视频序列的推进，包含OFA组件的模型的平均Alpha误差呈现出明显的下降趋势，并持续稳定在较低水平，而没有OFA组件的模型误差则相对较高且波动较大。
As observed in Figure~\ref{fig:sup_ofa_effect}, models calibrated with the OFA component (red curve) and without it (green curve) exhibit similar Alpha errors in the initial few frames. However, as the video sequence progresses, the model incorporating the OFA component shows a distinct downward trend in average Alpha error, stabilizing at a consistently lower level. In contrast, the model without OFA maintains a relatively higher error profile throughout the later frames.

% 这种现象清晰地揭示了OFA组件的有效性。由于我们的OFA损失 $\mathcal{L}_{\text{OFA}}$ 是从视频的第二帧开始计算并作用于优化过程的，它利用光流提供的时序先验信息来引导PTQ校准。这种引导作用不仅直接促进了模型学习更具时间连贯性的表征，从而减少了后续帧的预测误差和不稳定性，也间接作为一个有效的正则化项，帮助模型在整体上达到更高的抠图精度。
This phenomenon clearly demonstrates the effectiveness of the OFA component. Since our OFA loss, $\mathcal{L}_{\text{OFA}}$, is computed and applied to the optimization process starting from the second frame of a video, it leverages temporal prior information provided by optical flow to guide the PTQ calibration. This guidance not only directly encourages the model to learn more temporally coherent representations, thereby reducing prediction errors and instability in subsequent frames, but also indirectly acts as an effective regularizer, aiding the model in achieving higher overall matting accuracy.

\subsection{Experimental setup details}
% D.1 校准集构建
\paragraph{Calibration Set Construction}
% 如主论文中所述，我们的校准集非常小。具体而言，我们从VM视频数据集中选取了前64个视频片段。对于每个选定的视频片段，我们均匀采样其第0、2、4、6帧，总计构成256张图像作为校准数据。
As mentioned in the main paper, our calibration set is very small. Specifically, we selected the first 64 video clips from the VM video dataset. For each selected clip, we uniformly sampled frames at indices [0, 2, 4, 6], resulting in a total of $64 \times 4 = 256$ images for calibration.

% D.2 优化参数设置
\paragraph{Optimization Parameter Settings}
% 我们两阶段PTQ框架的优化参数设置如下：
The optimization parameters for our two-stage PTQ framework are set as follows:

% \begin{itemize}
%     \item \textbf{阶段一 (BIQ - 逐块初始量化)}：在此阶段，对于每个块的优化，我们使用Adam优化器，学习率固定为 $4 \times 10^{-5}$。每个块的优化迭代次数 (iters) 设置为20,000次。
%     \item \textbf{阶段二 (全局校准 - GAC 与 OFA)}：在此阶段，我们对所有可学习的校准参数（包括GAC的仿射变换参数 $\{\gamma_i, \beta_i\}$ 以及激活尺度因子 $\{s'_{a,i}\}$，和OFA损失的权重）进行联合优化。我们同样使用Adam优化器，学习率统一设置为 $1 \times 10^{-4}$。整个校准过程训练50个周期 (epochs)。对于光流辅助损失项 $\mathcal{L}_{\text{OFA}}$ 的权重因子 $\lambda$，我们将其设置为0.05。
% \end{itemize}
\begin{itemize}
    \item \textbf{Stage 1 (BIQ - Block-wise Initial Quantization)} During this stage, for the optimization of each block, we employ the Adam optimizer with a fixed learning rate of $4 \times 10^{-5}$. The number of optimization iterations for each block is set to 20,000.
    \item \textbf{Stage 2 (GAC and OFA)} In this stage, we jointly optimize all learnable calibration parameters, which include the affine transformation parameters $\{\gamma_i, \beta_i\}$ for GAC, the activation scaling factors $\{s'_{a,i}\}$, and implicitly the influence of the OFA loss. The Adam optimizer is used with a unified learning rate of $1 \times 10^{-4}$. The entire calibration process is run for 50 epochs. The weighting factor $\lambda$ for the Optical Flow-Assisted loss term ($\mathcal{L}_{\text{OFA}}$) is set to 0.05.
\end{itemize}

% D.3 硬件平台
\paragraph{Hardware Platform}
% 所有实验，包括模型量化、校准以及性能评估，均在单张NVIDIA RTX 4090 GPU (配备24GB显存) 上完成。
All experiments, including model quantization, calibration, and performance evaluation, were conducted on a single NVIDIA RTX 4090 GPU equipped with 24GB of VRAM.
% 值得一提的是，我们的整个PTQ校准流程对计算资源的需求较低，特别是显存占用，这使其能够很好地适应典型的PTQ任务场景，即在有限的资源下对预训练模型进行高效量化。
It is worth noting that our entire PTQ calibration pipeline has low computational resource requirements, especially in terms of VRAM usage, making it well-suited for typical video matting task scenarios where pre-trained models are efficiently quantized under limited resources.

\end{document}